\documentclass[acmtog]{acmart}

\usepackage{booktabs} 

\usepackage[ruled]{algorithm2e} 

\SetAlFnt{\small}
\SetAlCapFnt{\small}
\SetAlCapNameFnt{\small}
\SetAlCapHSkip{0pt}

\usepackage{multirow} 
\usepackage{bm} 

\newif\ifcameraready
\camerareadyfalse

\newcommand{\fref}[1]{Figure \ref{#1}}

\newcommand{\sref}[1]{Section \ref{#1}}  

\newcommand{\tref}[1]{Table \ref{#1}}

\newcommand{\aref}[1]{Appendix \ref{#1}}
\newcommand{\algref}[1]{Algorithm \ref{#1}}

\newcommand{\eref}[1]{Eq. \ref{#1}}

\newcommand{\error}{noise\xspace}

\colorlet{myblue}{blue}

\copyrightyear{2025}
\acmYear{2025}
\setcopyright{cc}
\setcctype{by}
\acmConference[SIGGRAPH Conference Papers '25]{Special Interest Group on Computer Graphics and Interactive Techniques Conference Conference Papers }{August 10--14, 2025}{Vancouver, BC, Canada}
\acmBooktitle{Special Interest Group on Computer Graphics and Interactive Techniques Conference Conference Papers (SIGGRAPH Conference Papers '25), August 10--14, 2025, Vancouver, BC, Canada}
\acmDOI{10.1145/3721238.3730678}
\acmISBN{979-8-4007-1540-2/2025/08}

\begin{document}
\title{Compensating Spatiotemporally Inconsistent Observations for Online Dynamic 3D Gaussian Splatting}

\author{Youngsik Yun}
\orcid{0000-0003-4398-7856}
\affiliation{%
 \institution{Yonsei University}
 \city{Seoul}
 \country{Republic of Korea}}
\email{bbangsik@yonsei.ac.kr}
 
\author{Jeongmin Bae}
\orcid{0009-0009-3376-2275}
\affiliation{%
 \institution{Yonsei University}
 \city{Seoul}
 \country{Republic of Korea}}
\email{jaymin.bae@yonsei.ac.kr}
 
\author{Hyunseung Son}
\orcid{0009-0009-1239-0492}
\affiliation{%
 \institution{Yonsei University}
 \city{Seoul}
 \country{Republic of Korea}}
\email{ghfod0917@yonsei.ac.kr}
 
\author{Seoha Kim}
\orcid{0009-0006-7456-701X}
\affiliation{%
\institution{Electronics and Telecommunications Research Institute}
\city{Daejeon}
\country{Republic of Korea}}
\email{hailey07@yonsei.ac.kr}

\author{Hahyun Lee}
\orcid{0000-0001-7043-7564}
\affiliation{%
\institution{Electronics and Telecommunications Research Institute}
\city{Daejeon}
\country{Republic of Korea}}
\email{hanilee@etri.re.kr}

\author{Gun Bang}
\orcid{0000-0003-4355-599X}
\affiliation{%
\institution{Electronics and Telecommunications Research Institute}
\city{Daejeon}
\country{Republic of Korea}}
\email{gbang@etri.re.kr}

\author{Youngjung Uh}
\authornote{Corresponding author.}
\orcid{0000-0001-8173-3334}
\affiliation{%
 \institution{Yonsei University}
 \city{Seoul}
 \country{Republic of Korea}}
\email{yj.uh@yonsei.ac.kr}

\renewcommand\shortauthors{Y. Yun et al}

\begin{abstract}
Online reconstruction of dynamic scenes is significant as it enables learning scenes from live-streaming video inputs, while existing offline dynamic reconstruction methods rely on recorded video inputs.
However, previous online reconstruction approaches have primarily focused on efficiency and rendering quality, overlooking the temporal consistency of their results, which often contain noticeable artifacts in static regions. 
This paper identifies that errors such as noise in real-world recordings affect temporal inconsistency in online reconstruction.
We propose a method that enhances temporal consistency in online reconstruction from observations with temporal inconsistency which is inevitable in cameras.
We show that our method restores the ideal observation by subtracting the learned error.
We demonstrate that applying our method to various baselines significantly enhances both temporal consistency and rendering quality across datasets. 
Code, video results, and checkpoints are available at \url{https://bbangsik13.github.io/OR2}.
\end{abstract}

%
%

%
%


\begin{teaserfigure}
  \includegraphics[width=\linewidth]{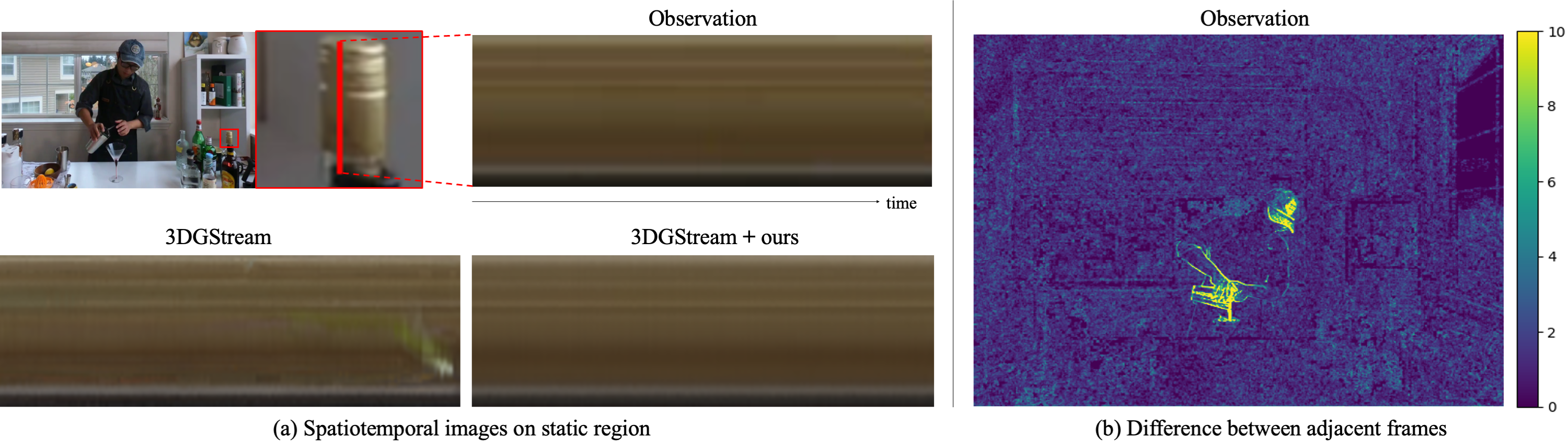}
  \caption{\textbf{Overview.} (a) Previous online reconstruction methods are temporally inconsistent. To visualize it, we horizontally concatenate the fixed vertical line over time, creating spatiotemporal images. 
  The previous method produces a noisy spatiotemporal image with artifacts in static regions, indicating temporal inconsistency. Our method enhances the temporal consistency of online reconstruction, producing a clearer spatiotemporal image.
(b) The spatiotemporally inconsistent observations cause the temporal inconsistency. All real images contain errors such as noise and these errors vary across camera views and video frames. We visualize a heatmap measuring the difference between subsequent observation frames, indicating that the static regions are temporally inconsistent by the error. Most errors have values below 4 in 8-bit images, and these subtle, nearly imperceptible values affect online reconstruction methods, leading to temporal inconsistent results. To address this issue, we propose to restore the ideal scene by disregarding the errors.
}
\Description[TeaserFigure]{TeaserFigure}
\label{fig:overview}
\end{teaserfigure}

\maketitle

\ifcameraready
    
\section{Introduction}
\label{sec:intro}
Reconstructing dynamic scenes from multi-view videos is a crucial problem in computer vision and graphics
to enable freely exploring novel viewpoints and timestamps. This capability has substantial potential for advancing applications in VR, AR, and XR.
Neural Radiance Fields (NeRFs)~\cite{mildenhall2021nerf} have significantly improved the fidelity of 3D reconstruction, and have been adopted to dynamic scenes~\cite{li2022neural}.
However, the relatively slow rendering speed of NeRFs hinders their use in real-time applications.
Recent 3D Gaussian splatting (3DGS)~\cite{kerbl3Dgaussians} has advantages in rapid training and real-time rendering, making it a mainstream approach for dynamic scene reconstruction~\cite{yang2023gs4d,wu20234dgaussians,duan20244drotor}.

However, most dynamic scene reconstruction methods are limited to offline reconstruction, which 
relies on full access to video sequences.
In addition, their results are typically not streamable, which cannot transmit part of the model of the specific moment to the user. Accordingly, they are incompatible with live-streaming applications, where 
only one frame is observable at a moment and the sent frames cannot be updated.
Furthermore, the offline methods face out-of-memory issues with long videos and depend on hyperparameters specific to different lengths.

To reconstruct dynamic scenes from streaming video inputs and enable the streaming of a learned model, recent approaches reconstruct dynamic scenes in online configuration by consecutively optimizing a model which reconstructs a moment~\cite{li2022streaming,sun20243dgstream}.
When the length of the video sequences is undefined, the online methods offer advantages over offline methods due to its unbounded nature~\cite{Wang2023rerf}.
Furthermore, online methods do not rely on length-related hyperparameters.
However, online reconstruction lacks temporal consistency in static regions, as shown in \fref{fig:overview}a. 
We aim to produce temporally consistent and high quality results by finding the problem and solution.

First, we find that the observations bear unnoticeable error rather than capturing the ideal signal as shown in \fref{fig:overview}b. Although we do not know the ideal signal, the colors in different moments at static regions should be the same, while the differences between adjacent frames are non-zero even at static regions. These errors are inevitable due to sensor noises or other reasons. By nature, they vary over time.
We suggest that the online reconstruction produces temporally inconsistent results at static regions because it observes a frame at a moment and overfits to the time-varying errors.
We demonstrate that the errors in the observation harm the temporal consistency through experiments on the synthetic dataset
where we can prepare the observations with and without errors (\sref{sec:method:problem}).

To this end, we propose \textit{observation-restoring online reconstruction} to reconstruct the ideal scene from imperfect observations with errors. 
The key idea is to reconstruct observations combining the Gaussians rendered on the images and residual maps which separately model the errors. We defer the details to \sref{sec:method:decomposing}.
Combined with baselines, our method improves both temporal consistency and the quality of rendered results. The experiments cover various baselines and datasets.
Additionally, our method reduces the number of Gaussians in the reconstruction leading to faster training and rendering, and lower memory footprint. Lastly, the performance measures have lower variance across multiple runs of our method compared to the baselines, indicating stability of our method.


\section{Related Work}
\label{sec:related}
In this section, we briefly review the literature on dynamic scene reconstruction using radiance fields, categorizing approaches into offline (\sref{sec:related:offline}) and online reconstruction (\sref{sec:related:online}). 
Then, we discuss existing works reconstructing the radiance field from inaccurate observations (\sref{sec:related:correction}).

\subsection{Offline Reconstruction of Radiance Fields}
\label{sec:related:offline}
To extend NeRFs~\cite{mildenhall2021nerf} to reconstruct dynamic scenes, several approaches deform rays~\cite{pumarola2020d,park2021nerfies,park2021hypernerf} or expand 3D representations to 4D~\cite{fridovich2023k,cao2023hexplane}, focusing on improving rendering performance, memory efficiency, and training speed.
NeRFPlayer~\cite{song2023nerfplayer} enables the streaming of a learned model by representing each time of the scene with local feature channels.

The emergence of real-time rendering in 3D Gaussian Splatting (3DGS)~\cite{kerbl3Dgaussians} encourages extensive research in primitives-based dynamic scene reconstruction. Similar to Dynamic NeRFs, several works reconstruct dynamic scenes by deforming the Gaussians~\cite{wu20234dgaussians,bae2024ed3dgs,li2024st4dgs} or extending 3D Gaussians to 4D Gaussians~\cite{yang2023gs4d,cho20244dscaffold}. 

Moreover, due to memory constraints, HyperReel~\cite{attal2023hyperreel} and STG~\cite{li2023spacetimegaussians} independently reconstruct each video segment from a single video sequence. This shows visual inconsistencies between the reconstructed results.
SWinGS~\cite{shaw2024swings} segments videos into size-varying segments based on motion intensity, sharing one frame between adjacent segments and training models for each segment. These models are subsequently fine-tuned across total frames to enhance temporal consistency. 
 
However, these offline reconstruction methods cannot process live-streaming input, relying on recorded video input. Moreover, previous temporal consistency enhancement methods require full-frame fine-tuning which is unsuitable for online reconstruction.

\begin{figure*}[tb!]
    \centering
    \includegraphics[width=0.88\linewidth]{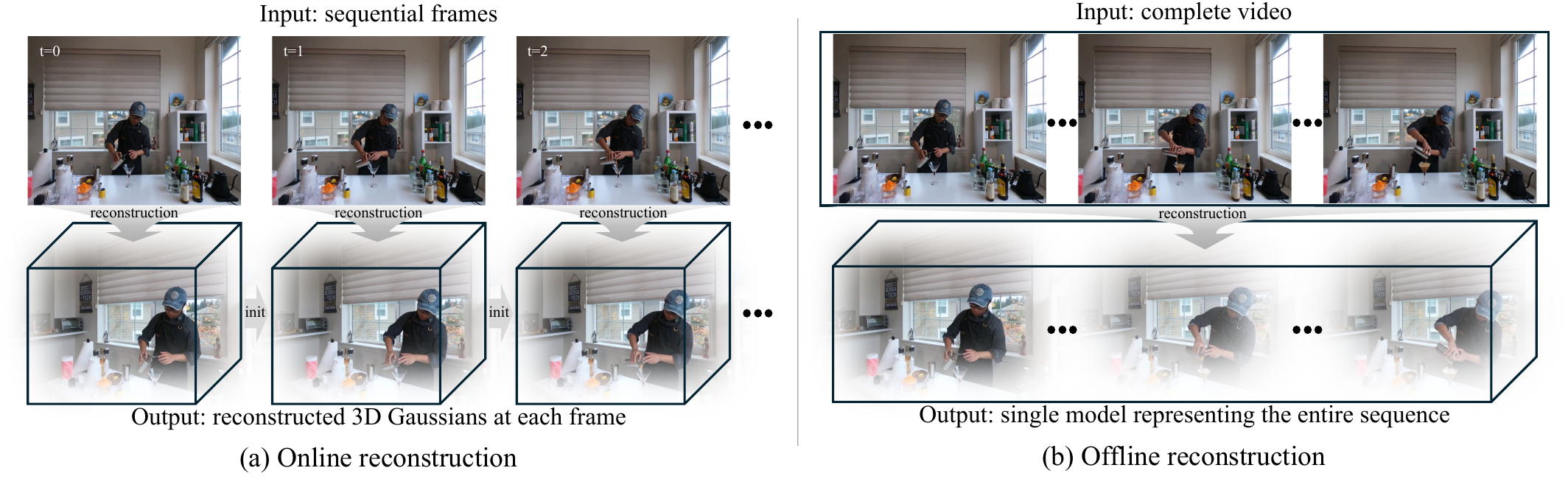}
    \caption{
    \textbf{Comparison of Online and Offline Reconstruction.}
    (a) Online reconstruction takes streaming video as input and sequentially learns a model for each frame.
(b) In contrast, offline reconstruction takes the complete set of video frames as input and produces a single model representing the entire sequence.}
    \label{fig:setting}
\Description[Figure]{Figure of the overall framework}
\end{figure*}

\subsection{Online Reconstruction of Radiance Fields}
\label{sec:related:online}
Online reconstruction methods sequentially optimize the scene at a given time, enabling live-streaming input processing and learned model streaming.
The simplest way for online reconstruction is to train a model from each frame of live-streaming input. However, as most video contents contain significant redundancy across frames, reconstructing each frame independently is inefficient. 
Therefore, StreamRF~\cite{li2022streaming} and ReRF~\cite{Wang2023rerf} consecutively learn the differences from previous frame for efficiency. 

Recent research on online reconstruction predominantly uses 3D Gaussian Splatting (3DGS) as a representation for real-time rendering. These methods optimize the difference between consecutive frame models~\cite{luiten2023dynamic}. To enhance training efficiency, 3DGStream~\cite{sun20243dgstream} employs iNGP~\cite{mueller2022instant}, an implicit network for optimizing the residual instead of directly optimizing Gaussian parameters. HiCoM~\cite{gao2024hicom} introduces an explicit hierarchical motion field to reduce model capacity and utilizes a reference frame prediction to accelerate training. 

However, these studies focus on the efficiency and visual quality of each frame, overlooking the temporal consistency of their results. We aim to enhance the temporal consistency of online reconstruction methods.

\subsection{Optimizing Radiance Field from Imperfect Data}
\label{sec:related:correction}
NeRF$--$~\cite{wang2021nerfmm}, BARF~\cite{lin2021barf}, and SCNeRF~\cite{SCNeRF2021} jointly optimize NeRF and camera parameters, reducing the need for known camera parameters in static scene reconstruction. NoPe-NeRF~\cite{bian2023nopenerf}, CF-3DGS~\cite{fu2024cf3dgs} leverage depth priors for more accurate pose estimation. RoDyNeRF~\cite{liu2023robust} extends this approach by jointly optimizing dynamic NeRF and camera parameters to correct inaccurate camera poses. 
SyncNeRF~\cite{Kim2024Sync} jointly optimizes dynamic radiance fields and time offsets to reconstruct high-quality 4D models from unsynchronized videos.

DeblurNeRF~\cite{li2022deblurnerf}, Deblur3DGS~\cite{lee2024deblurring}, Robust3DGaussians~\cite{darmon2024robust}, and BAD-Gaussians~\cite{zhao2024badgaussians} learn blur formation to obtain sharp reconstruction from blurry inputs. Deblur4DGS~\cite{wu2024deblur4dgs} extends this concept to learning blur formation for sharp dynamic scene reconstruction from blurry monocular videos.
DehazeNeRF~\cite{chen2023dehazenerf}, SeaSplat~\cite{yang2024seasplat}, and DehazeGS~\cite{yu2025dehazegs} reconstruct a clean scene from hazy inputs.
NeRF-W~\cite{martin2021nerf}, WildGaussians~\cite{kulhanek2024wildgaussians}, and NeRF-On-the-go~\cite{Ren2024NeRF} reconstruct 3D scenes from internet images by learning to ignore distractors, such as moving objects.

Sharing a key insight with our method, RawNeRF~\cite{mildenhall2021rawnerf} begins with the premise that all real images contain noise.
RawNeRF demonstrates that NeRF can remove zero mean noise in the observation by averaging information using multi-view correspondence in high dynamic range (HDR) space. However, converting an HDR image where per-pixel noise follows zero-mean Gaussian distribution into a low dynamic range (LDR) image through nonlinear mapping alters the noise distribution, corrupting the noise distribution to a nonzero mean distribution. As a result, noise removal becomes less effective by averaging information in LDR images.

Several methods have attempted to reconstruct from inaccurate inputs, but they have generally focused on modeling errors under specific settings. 
In contrast, we address errors that occur naturally in real-world recording scenarios, where the error factors are diverse and ambiguous, making it challenging to model each factor separately.


\section{Observation-restoring online reconstruction}
\label{sec:method}
In this section, we first provide a preliminary of the online reconstruction of dynamic 3D Gaussian Splatting (\sref{sec:method:prelim}). Next, we introduce our problem setting (\sref{sec:method:problem}). Finally, we propose its solution (\sref{sec:method}).
 Algorithm 1 in the Appendix summarizes our method.

\subsection{Preliminary: Online Dynamic 3D Gaussian Splatting}
\label{sec:method:prelim}

Offline reconstruction aims to optimize a single model representing the entire sequence from the entire duration of multi-view videos. 
In contrast, online reconstruction aims to reconstruct a model of a \textit{moment} and update the model to sequentially reconstruct consecutive frames, for given sequentially captured multi-view images. This configuration is inevitable for live streaming free-viewpoint videos.
Noting that the offline approaches assume a fixed duration for a model, the online configuration is advantageous for training long and arbitrary-length videos because the length of reconstruction increases while capturing a scene. \fref{fig:setting} conceptually compares online and offline reconstruction.

We provide a formal configuration.
Let $G_t$ be a set of Gaussians at time $t$, defined as $G_t = (\boldsymbol{\mu}_t,\mathbf{q}_t,\mathbf{s}_t,\boldsymbol{\sigma}_t,\mathbf{Y}_t)$ where
$\boldsymbol{\mu}, \mathbf{q}, \mathbf{s}, \boldsymbol{\sigma}$, and $\mathbf{Y}$ denotes mean, rotation, scale, opacity, and spherical harmonics (SH) coefficients of the Gaussians, respectively\footnote{For brevity, we omit the indices of each Gaussian and its parameters.}.
Given a set of cameras $\mathbb{V}$, let $I_t^v$ be the observation of a scene from a camera $v \in \mathbb{V}$ at time $t \in \mathbb{Z}_{\geq 0}$\footnote{$t$ is not bounded above in contrast to offline configuration.}. 

The attributes of the Gaussians $G_0$ are initialized from Structure-from-Motion (SfM) points.
These attributes are optimized by rendering $G_0$ across multiple cameras $v$, minimizing the loss between the rendered image $\hat{I}_0^v$ and their corresponding observation $I_0^v$. 
The number of Gaussians is adaptively adjusted by densification and pruning during optimization. Gaussians are split and cloned if their pixel-space gradient\footnote{Gradient of the loss with respect to the coordinates projected on the image planes} exceeds the predefined threshold, and pruned if their opacity is low. The Gaussians $G_{t+1}$ at the next frame are obtained by deforming the attributes of the optimized $G_t$ from the previous frame.

By design, deforming the Gaussians from previous frames cannot reconstruct new objects.
Therefore, 3DGStream~\cite{sun20243dgstream} introduces new Gaussians $G_t^{\text{new}}$ to represent novel objects.
However, 3DGStream uses these new Gaussians for a single frame and discard them, leading to temporally inconsistent results in dynamic regions (Appendix D.2).
To solve this, similar to previous works~\cite{gao2024hicom,girish2024queen}, we propagate both new and deformed Gaussians to subsequent frames to enhance temporal consistency.

\begin{figure}[tb!]
    \centering
    \includegraphics[width=\linewidth]{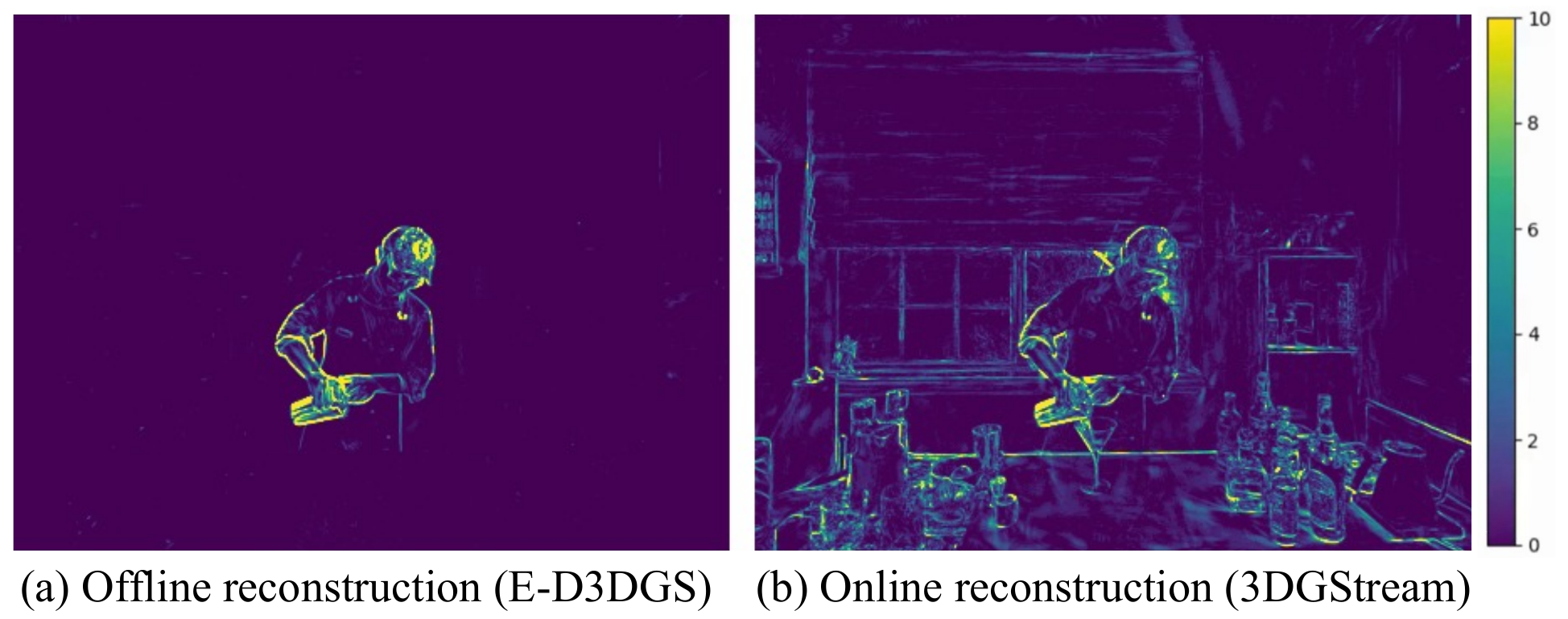}
    \caption{\textbf{Differences between Consecutive Frames.} 
    (a) While the offline reconstruction method maintains temporal consistency in static regions, 
    (b) the online reconstruction method lacks temporal consistency.}
    \label{fig:offline_vs_online}
\Description[Figure]{Different results between offline reconstruction and online reconstruction methods.}
\end{figure}

\begin{figure}[tb!]
    \centering
    \includegraphics[width=\linewidth]{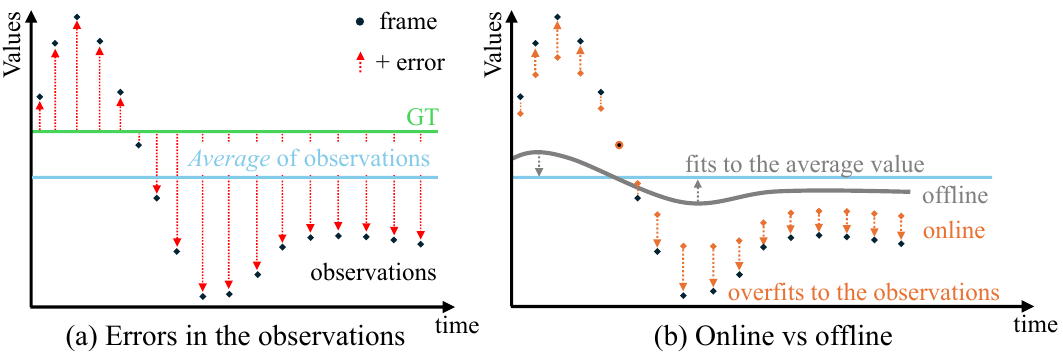}
    \caption{\textbf{Conceptual Figure of the Problem Statement} (a) Each observation value contains errors added to the ground truth. (b) Due to limited data accessibility, online reconstruction tends to overfit each observation, whereas offline reconstruction converges to the global average over time.}
    \label{fig:concept}
\Description[Figure]{The concept of observation errors and the differences between offline and online reconstruction methods.}
\end{figure}

\begin{figure}[tb!]
    \centering
    \includegraphics[width=0.95\linewidth]{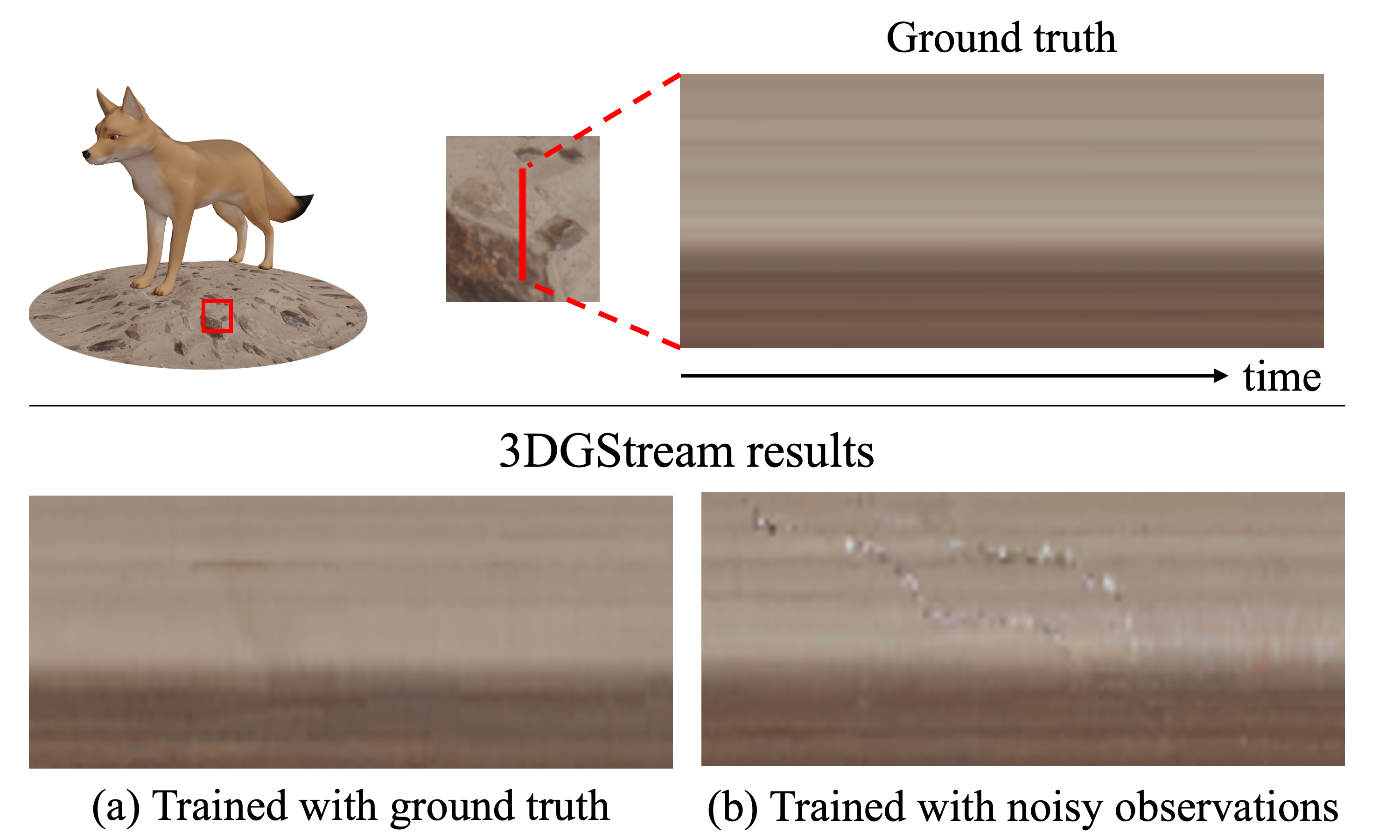}
    \caption{
    \textbf{Comparison of Spatiotemporal Images under Two Settings.}
(a) Training with ground truth images produces clearer results than (b) training with noisy observations created by adding noise to the ground truth. This demonstrates that noise harms the temporal consistency. \copyright DuDeHTM \href{https://creativecommons.org/licenses/by-nc/4.0}{(CC BY-NC 4.0)}
    }
    \label{fig:effect_of_error}
\Description[Figure]{Effect of \error in online reconstruction.}
\end{figure}

\begin{figure}[tb!]
    \centering
    \includegraphics[width=0.95\linewidth]{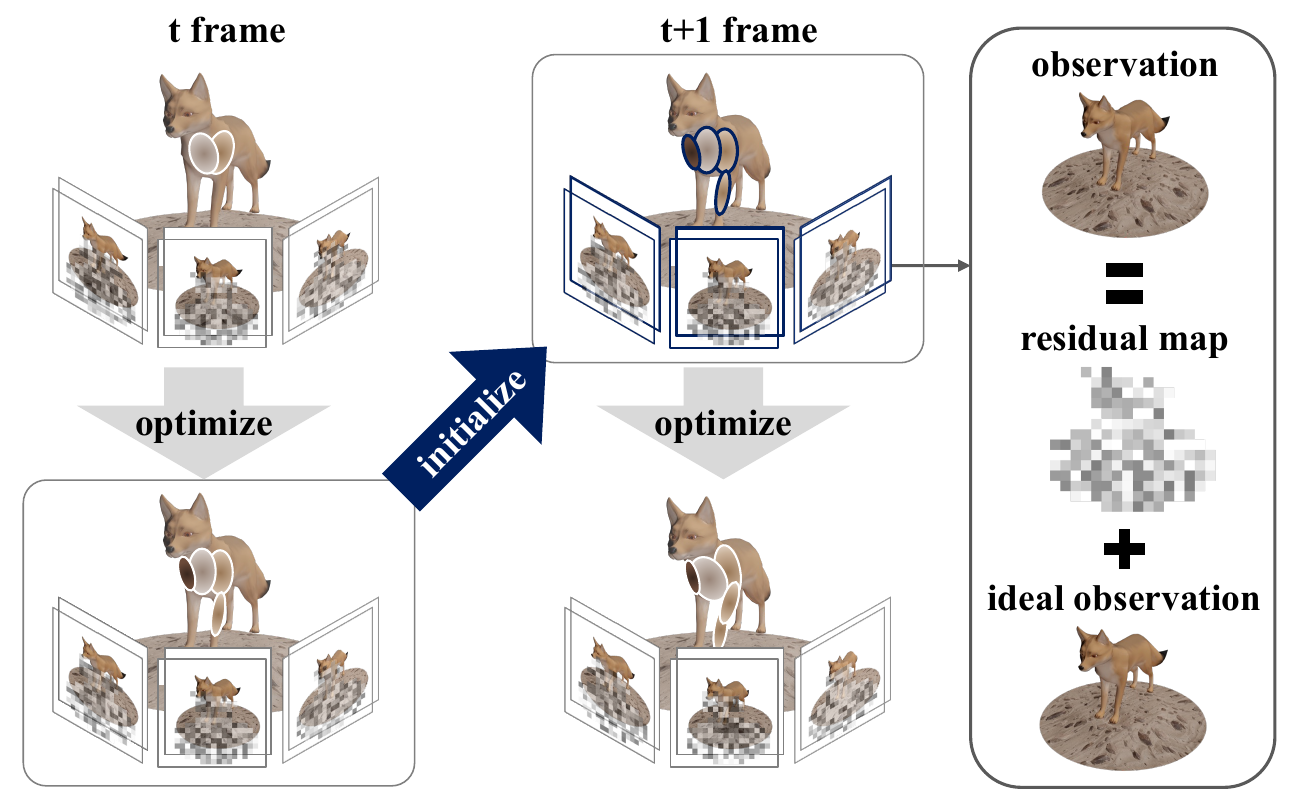}
    \caption{\textbf{Framework.} We jointly optimize 3D Gaussians and residual maps for each camera view. The residual maps model observation errors. For subsequent frames, both Gaussians and residual maps are initialized with values from the previous frame. We model an observation as a combination of the ideal observation and residual map. \copyright DuDeHTM \href{https://creativecommons.org/licenses/by-nc/4.0}{(CC BY-NC 4.0)} 
    }
    \label{fig:framework}
\Description[Figure]{Figure of the overall framework}
\end{figure}

\subsection{Problem Statement}
\label{sec:method:problem}

\begin{figure*}[tb!]
    \centering
    \includegraphics[width=\linewidth]{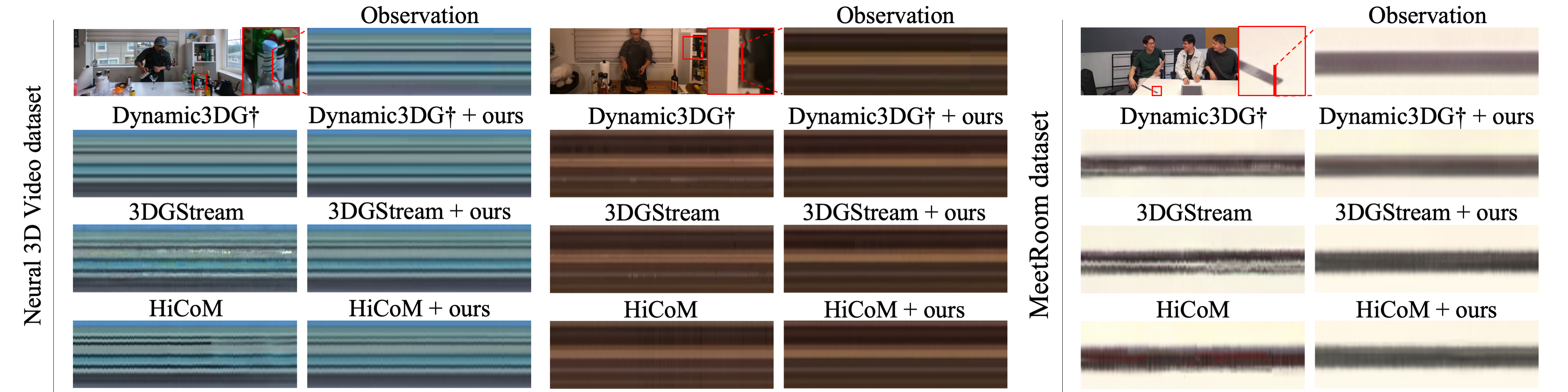}
    \caption{
    \textbf{Qualitative Comparison on Neural 3D Video and MeetRoom Dataset.} Our method produces more temporal consistent spatiotemporal images.}
    \label{fig:qual}
\Description[Figure]{qualitative results of the N3V and MeetRoom dataset}
\end{figure*}

Online reconstruction methods lack temporal consistency whereas offline reconstruction methods do not, as shown in \fref{fig:offline_vs_online}. Intriguingly, the observations lacks temporal consistency between adjacent frames even in static regions as shown in \fref{fig:overview}b. We hypothesize that the observations are corrupted and do not ideally observe the original scene.

Offline reconstruction does not suffer from the above problem because the model converges to the average of all frames because the model with limited capacity cannot fit all tiny differences between the frames as shown in \fref{fig:concept}.
This averaging phenomenon ignores temporally varying errors in the observations, resulting in relatively consistent values for static regions.
In contrast, online reconstruction converges to render a specific frame because it can access only the single target frame.
As a result, the reconstructed result overfits the temporally varying errors in its corresponding observations, leading to temporally inconsistent reconstruction even in static regions.

We support the problem statement with a toy experiment. We reconstruct a synthetic dynamic 3D asset for given observations prepared by rendering it on fixed viewpoints, with and without noise\footnote{We add Gaussian and Poisson noise.}. While the results from ground truth observations are mostly clear (\fref{fig:effect_of_error}a), the results from noisy observation exhibit flickering artifacts (\fref{fig:effect_of_error}b).

We formulate the observations $\tilde{I}_t^v$ capturing the scene as a combination of ideal observations $I_t^v$ and errors $M_t^v$, and propose to restore the ideal scene with temporal consistency by disregarding the errors.

\subsection{Decomposing Errors in the Observations}
\label{sec:method:decomposing}
\paragraph{Learnable residual map}
\label{sec:method:residual}
To address observations containing errors that vary across camera views $v$ and frames $t$, we separately model these errors with additional residual maps $\hat{M}_t^v \in \mathbb{R}^{3 \times H \times W}$, which are optimized as learnable parameters during training:
\begin{equation}
\label{eq:separate}
    \tilde{I_t}=I_t+M_t=\hat{I}_t+\hat{M}_t,    
\end{equation}
where $\hat{I}_t$ and $\hat{M}_t$ are rendered image and estimated residual, respectively, and we omit $v$ for brevity. $H$ and $W$ denote the height and width of the image $I_t$.
During optimization, we minimize the difference between $\tilde{I_t}$ and $\hat{I}_t+\hat{M}_t$ by jointly updating $G_t$ and $\hat{M}_t$. An optimization step becomes:
\begin{equation}
    G_t, \hat{M_t} \gets \text{Adam}(\nabla L_\text{total}(\tilde{I}_t,\hat{I}_t+\hat{M_t})),
\label{eq:backward}
\end{equation}
where $L_\text{total}$ will be described later.

Separately modeling the residual maps restores the ideal observations as follows.
In \eref{eq:separate}, we expect the rendered images and the estimated residual maps to match the corrupted observation together. 
Fitting multi-view inconsistent high-frequency noise with a residual maps are easier than optimizing one set of Gaussians. Moreover, learning multi-view and temporal inconsistent noise with a residual map is easier than fine-tuning a well-trained Gaussians from the previous frame.
As a result, the optimized Gaussians reconstruct the true scene without corruption. I.e., the rendered images become the ideal observation without errors.

\paragraph{First Frame Reconstruction}
\label{sec:method:init}
To reconstruct the first frame, we initialize Gaussians from SfM points~\cite{schonberger2016colmap} and residual maps to zero.
We then jointly optimize them same as 3DGS~\cite{kerbl3Dgaussians}. To prevent the residual maps from dominating the reconstruction, they are frozen at zero until the Gaussians start densification.

In addition, we apply L1 regularization on opacity~\cite{kheradmand2024gsmcmc,bae2024ed3dgs} as resetting opacity in 3DGS~\cite{kerbl3Dgaussians} disrupts stable optimization of the residual maps: 
$L_\text{opa} = \sum_i{ \|\sigma_t^i\|_1 }$,
where $\sigma_t^i$ denotes opacity of the $i$-th Gaussian at time $t$.
We also employ L1 regularization on the residual map to prevent overfitting to view-conditioned color:
$L_\text{res} = \|\hat{M}_t^v\|_1.$

The total loss for the first frame is:
\begin{equation}
    L_\text{total} = (1 - \lambda)L_1 + \lambda L_\text{D-SSIM} + \lambda_\text{opa} L_\text{opa} + \lambda_\text{res} L_\text{res}.
\label{eq:loss:first}
\end{equation}

\paragraph{Sequential Frame Reconstruction}
\label{sec:method:per-frame}
We consecutively propagate new Gaussians across subsequent frames: $G_{t+1} \gets (G_t,G_t^\text{new})$ because using the new Gaussians in only a single frame and discarding them harms the temporal consistency.
We do not use $L_\text{opa}$ in sequential frame reconstruction to ensure fair comparison with baselines.
Finally, the total loss for the remaining frames is:    
\begin{equation}
    L_\text{total} = (1 - \lambda)L_1 + \lambda L_\text{D-SSIM} + \lambda_\text{res} L_\text{res}.
\label{eq:loss:seq}
\end{equation}

\begin{table*}[tbh!]
    \centering
    \caption{\textbf{Average Performance on Neural 3D Video and MeetRoom Dataset}. Computational cost was measured on \texttt{flame\_salmon} with RTX A5000.}
    \resizebox{0.9\linewidth}{!}{
    \begin{tabular}{l|ccc|ccc|cc}
    \toprule
    \multirow{2}{*}{Model}  &   \multicolumn{3}{c|}{Neural 3D Video} &   \multicolumn{3}{c|}{MeetRoom} &   \multicolumn{2}{c}{Computational Cost}  \\               
    & PSNR $\uparrow$ & SSIM $\uparrow$ & mTV$_{\times100}$ $\downarrow$ & PSNR $\uparrow$ & SSIM $\uparrow$ & mTV$_{\times100}$ $\downarrow$ & \# Gaussians $\downarrow$ & Training time $\downarrow$ \\
    \midrule
    Dynamic3DG$\dagger$ & 32.48 & 0.960 & 0.153 & 30.79 & 0.952 & 0.175 & 315K & 136. sec/frame         \\ 
    Dynamic3DG$\dagger$ + ours  & \textbf{33.03} & \textbf{0.961} & \textbf{0.053} & \textbf{31.29} & \textbf{0.953} & \textbf{0.126} & \textbf{246K} & \textbf{130.} sec/frame         \\ \midrule
    3DGStream           & 32.58 & 0.960 & 0.178 & 31.78 & 0.955 & 0.049 & 315K & 13.3 sec/frame         \\
    3DGStream + ours    & \textbf{33.13} & \textbf{0.961} & \textbf{0.103} & \textbf{32.29} & \textbf{0.957} & \textbf{0.024} & \textbf{268K} & \textbf{13.1} sec/frame        \\ \midrule
    HiCoM               & 32.09 & 0.956 & 0.164 & 29.42 & 0.936 & 0.127 & 307K & 11.0 sec/frame         \\
    HiCoM + ours        & \textbf{32.46} & \textbf{0.957} & \textbf{0.118} & \textbf{29.48} & \textbf{0.938} & \textbf{0.045} & \textbf{218K} & \textbf{10.2} sec/frame        \\ 
    \bottomrule
    \end{tabular}
    }
    \label{tab:quant}
\end{table*}

\section{Experiment}
\label{sec:exp}
In this section, we first provide the reasons for choosing the dataset, implementation, and the evaluation metrics (\sref{exp:criteria}). 
Next, we demonstrate our improvements in both temporal consistency and visual quality upon the baselines (\sref{sec:exp:comparison}).
We then show our proposed method effectively restores the observations (\sref{sec:exp:restoration}). 
Finally, we conduct an ablation study to assess the performance of our method (\sref{sec:exp:ablation}).


\subsection{Datasets, Implementation, and Metrics}
\label{exp:criteria}
\subsubsection{Datasets}
\label{exp:criteria:data}
Following the prior works~\cite{luiten2023dynamic,sun20243dgstream}, we use undistorted images and SfM points in our experiments.  Additional experiments on the original images and MVS points~\cite{wu20234dgaussians,gao2024hicom} are detailed in  Appendix C. We evaluated the baselines on these datasets.
\paragraph{Neural 3D Video~\cite{li2022neural}} The dataset contains six scenes captured by 17 to 21 multi-view cameras over 300 frames.
Following the prior works~\cite{wang2022mixed,fridovich2023k}, we downsample the video by a factor of two and remove the unsynchronized video in \texttt{coffee\_martini} scene.

\paragraph{MeetRoom~\cite{li2022streaming}} The dataset contians three scenes captured by 14 multi-view cameras over 300 frames.

\paragraph{Sync-NeRF~\cite{Kim2024Sync}} The dataset is rendered from dynamic 3D asset with static regions, providing a training dataset without errors (ground truth dataset). 
For our experiments, we use the modified \texttt{fox}\footnote{\copyright DuDeHTM \href{https://creativecommons.org/licenses/by-nc/4.0}{(CC BY-NC 4.0)} } scene (see  Appendix B.3 for details). We train with noisy observations\footnotemark[1] and evaluate against the ground truth.


\begin{figure}[tb!]
    \centering
    \includegraphics[width=\linewidth]{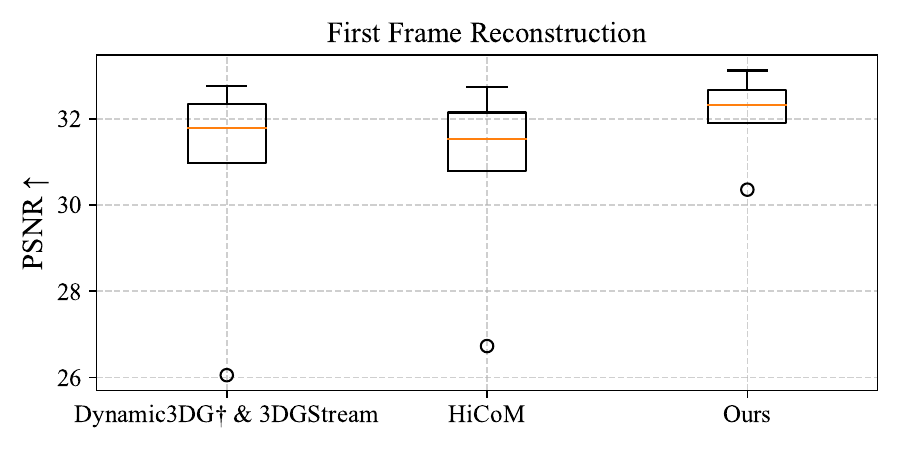}
    \caption{\textbf{First Frame Reconstruction}. Box plots comparing first frame reconstruction quality across 10 runs for each scene in the Neural 3D Video dataset.}
    \label{fig:initial_frame}
\Description[Figure]{Initial frame reconstruction}
\end{figure}

\subsubsection{Implementation}
\label{exp:criteria:imple}
We apply our method to the state-of-the-art online reconstruction methods (Dynamic3DG~\cite{luiten2023dynamic}, 3DGStream~\cite{sun20243dgstream}, and HiCoM~\cite{gao2024hicom}) and compare the performance with the original implementations. StreamRF \cite{li2022streaming} is excluded because it is hardly reproducible.

We follow the official implementations except for Dynamic3DG. Due to the poor results of Dynamic3DG on the datasets with forward-facing configuration, we modified it to improve performance, naming it Dynamic3DG$\dagger$ (see Appendix B.1 for details). To ensure a fair comparison, we apply our method to HiCoM using the same pruning strategy proposed in it\footnote{HiCoM maintains compactness by pruning the K lowest opacity Gaussians from the combined set of deformed and new Gaussians, where K equals the number of new Gaussians. However, this pruning strategy degrades visual quality.}.

As 3DGS~\cite{kerbl3Dgaussians} has randomness in optimization\footnote{Please refer to https://github.com/graphdeco-inria/gaussian-splatting/issues/89}, we select the 3D Gaussians with the highest PSNR from 10 runs as the initialization for reconstructing the sequential frame. This protocol is confirmed by correspondence with the authors of 3DGStream.
We use the same 3D Gaussians at the first frame for both Dynamic3DG$\dagger$ and 3DGStream\footnote{Note that once the 3D Gaussians at the first frame are fixed, sequential frames bear minimal variance, indicating that the randomness in online reconstruction primarily derives from the first frame.}.

In our experiments, we do not prune Gaussians with large radii in pixel-space because it degrades performance.
We set the SH degree to 3 when applying our method (See \sref{sec:exp:ablation}).


\subsubsection{Metrics}
\label{exp:criteria:metric}
We report the quality of rendered images using
Peak Signal-to-Noise Ratio (PSNR), Structural Similarity Index Measure (SSIM), and masked Total Variation (mTV).
PSNR quantifies pixel color error between the rendered video and the test dataset. 
We measure SSIM to account for the perceived similarity of the rendered image. Additionally, 
we measure mTV to measure temporal consistency in static regions indicated by predefined masks. Appendix B.3 provides details about the masks used in the experiments. Higher PSNR and SSIM values and lower mTV values indicate better visual quality. For readability, we multiply mTV by 100 and refer to it as mTV$_{\times100}$. 
 

\subsection{Comparison on Real-world Dataset}
\label{sec:exp:comparison}

\subsubsection{Visual Quality}
\label{sec:exp:comparisions:init}

\paragraph{The First Frame} Our method improves PSNR at the first frame in both median and highest values as shown in \fref{fig:initial_frame}. The box plot is drawn with 10 runs per scene in the Neural 3D Video dataset. In addition, our method achieve lower variance, implying higher stability than the baselines.

\paragraph{All Frames}
Consecutively applying our method to all frames improves both PSNR and SSIM, achieving superior visual quality as shown in \tref{tab:quant}. 

\begin{figure}[tb!]
    \centering
    \includegraphics[width=\linewidth]{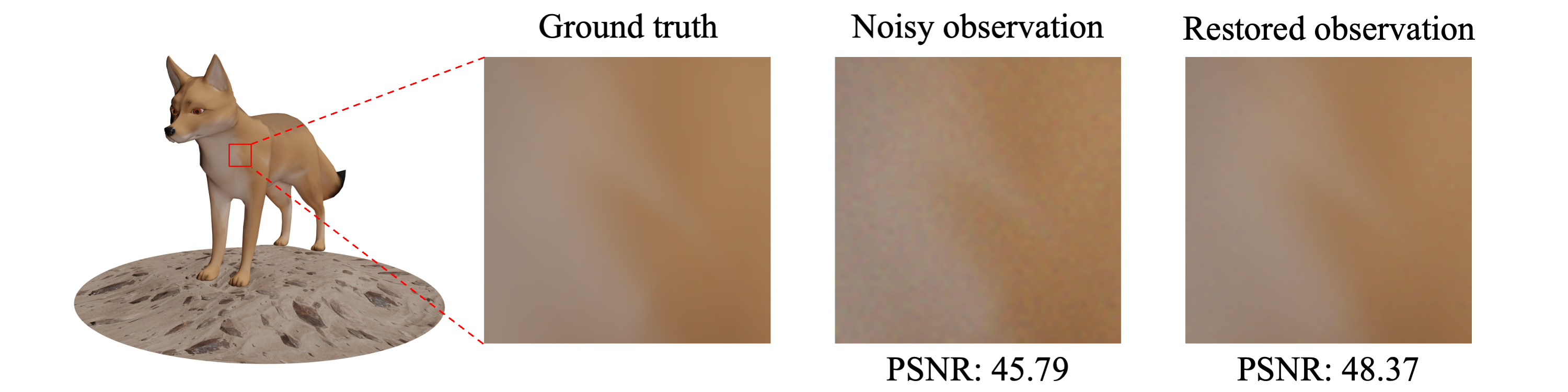}
    \caption{\textbf{Observation Restoration}. Our learned residual maps successfully restore the noisy observations. \copyright DuDeHTM \href{https://creativecommons.org/licenses/by-nc/4.0}{(CC BY-NC 4.0)}
    }
    \label{fig:restoration}
\Description[Figure]{Restoration.}
\end{figure}
\begin{table}[tb!]
    \centering
    \caption{\textbf{Average Performance on Sync-NeRF Dataset}.}
    \resizebox{\linewidth}{!}{
    \begin{tabular}{l|ccc}
        \toprule
        Model & PSNR $\uparrow$ & SSIM $\uparrow$ & mTV$_{\times100}$ $\downarrow$  \\
        \midrule
         3DGStream & 38.79 & 0.982 & 0.483 \\
         3DGStream on denoised observations & 38.63 & 0.979 & 0.417 \\
         3DGStream + ours & \textbf{39.16} & \textbf{0.987} & \textbf{0.318} \\
        \bottomrule
    \end{tabular}
    }
    \label{tab:synthetic}
\end{table}


\subsubsection{Temporal Consistency}
\fref{fig:qual} presents spatiotemporal images of static areas. While baseline methods show noisy results, combining our method produces clear outputs, indicating improved temporal consistency. 
In \tref{tab:quant}, applying our method reduces mTV, demonstrating enhanced temporal consistency.


\subsubsection{Computational Cost}
\label{sec:exp:comparisions:cost}
\tref{tab:quant} shows the computational costs.
Our approach reduces the number of Gaussians for each baselines. 
Although we optimize additional parameters, the reduced number of Gaussians decreases the training time.
Notably, the residual maps are only used during training and are not stored.


\subsection{Comparison on Synthetic Dataset}
\label{sec:exp:restoration}
\subsubsection{Restoration}
To demonstrate the ability of our learned residual map to restore the observations, we conduct experiments on Sync-NeRF dataset. 
We restore the train view observation to its ground truth by subtracting the residual map from the corresponding observation.
This restoration increases PSNR (+2.58) with clear visual quality, as shown in \fref{fig:restoration}.

\subsubsection{Comparison to Preprocessing}
We additionally compare our method against combination of 3DGStream and a denoising model \cite{liang2022vrt}, where 3DGStream reconstructs the scene from the denoised observations.

As shown in \tref{tab:synthetic}, our method outperforms competitors in both visual quality (PSNR) and temporal consistency (mTV). Using a denoising model also enhances temporal consistency, which supports our problem settings. Nonetheless, its effectiveness remains lower than ours\footnote{Moreover, \citet{liang2022vrt} degrade visual quality through over-smoothing, and the preprocessing time (196 seconds per frame) significantly exceeds the sequential frame reconstruction time (6 seconds per frame).}.

\begin{table}[tb!]
\centering
\caption{\textbf{Quantitative Comparison of Ablation Study.}}
\resizebox{0.85\linewidth}{!}{
\begin{tabular}{l|ccc}
\toprule
    Method & PSNR $\uparrow$ & SSIM $\uparrow$ & mTV$_{\times100}$ $\downarrow$ \\
\midrule
3DGStream + ours & \textbf{33.13} & 0.961 & \textbf{0.103} \\ \midrule
w/o residual maps & 32.65 & 0.960 & 0.161 \\
w/o SH degree 3 & 33.08 & 0.961 & 0.129 \\
w/o reuse new Gaussians & 33.06 & 0.960 & 0.118 \\
\bottomrule
\end{tabular}
}
\label{tab:ablation}
\end{table}
\begin{figure}[tb!]
    \centering
    \includegraphics[width=0.95\linewidth]{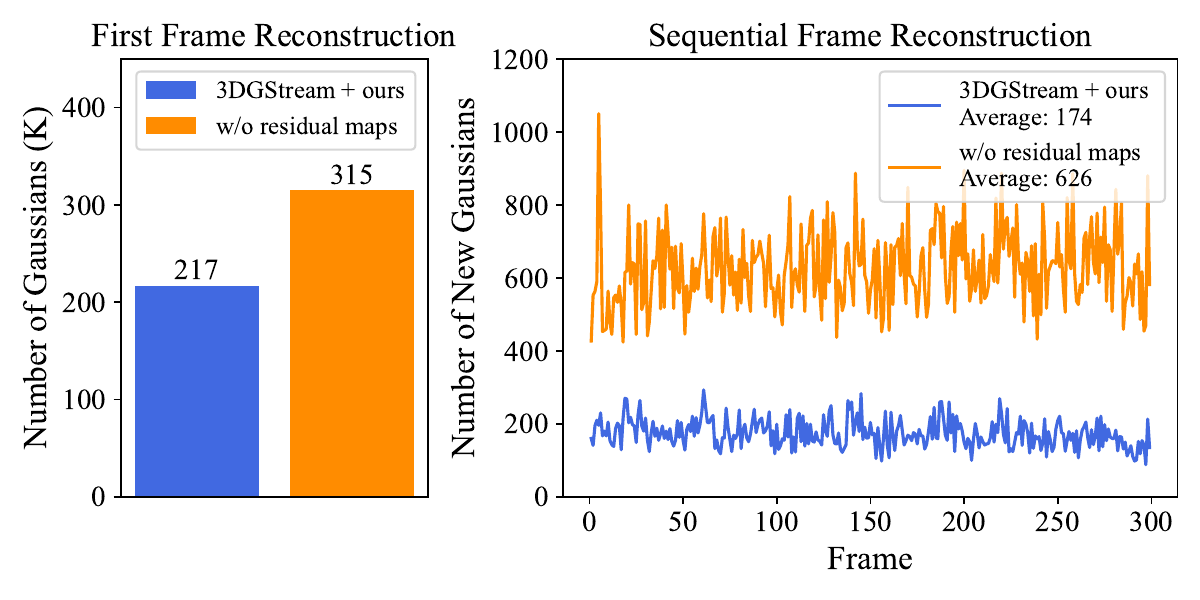}
    \caption{\textbf{Ablation Study of Residual Maps}. Our proposed residual maps reduces the number of Gaussians in both first frame reconstruction and subsequent new Gaussian additions, measured on the \texttt{flame\_salmon} scene.}
    \label{fig:abl_new_gaussian}
\Description[Figure]{spatiotemporal images}
\end{figure}


\subsection{Ablation Study}
\label{sec:exp:ablation}
\subsubsection{Learnable Residual Maps}
To validate the effect of the residual map, we compared ``3DGStream + ours'' to a version without the residual map, labeling ``w/o residual maps''.
As shown in Table 2, using residual maps improves both temporal consistency (reduced mTV) and visual quality (increased PSNR).
Our method reduces unnecessary Gaussians by separating observation errors that inflate the pixel-space gradient.
Leveraging our residual maps decreases the number of Gaussians, reducing $G_0$ to 0.69$\times$ and $G_t^{\text{new}}$ to 0.28$\times$, as shown in \fref{fig:abl_new_gaussian}.

\subsubsection{SH Degree}
We analyze the variation of reconstruction performance due to different SH degrees.
The higher the SH degrees, the higher the model capacity.
In the first frame reconstruction, 3DGS~\cite{kerbl3Dgaussians} with higher model capacity tends to overfit observation errors, performing better at SH degree 1, as shown in \fref{fig:abl_SH}. 
In contrast, our residual maps fit these errors instead of SH, allowing Gaussians to effectively use their higher model capacity without overfitting the errors. Moreover, higher 3D Gaussian capacity prevents residual maps from overfitting view-dependent color, thus achieving superior performance at SH degree 3. 
Consequently, in sequential frame reconstruction, setting the SH degree to 3 enhances visual quality and temporal consistency compared to lower SH degree 1, as demonstrated in \tref{tab:ablation} ``w/o SH degree 3''.

\subsubsection{New Gaussians}
Propagating new and deformed Gaussians to the next frame enhances visual quality and temporal consistency in static regions compared to using new Gaussians only in specific frames labeled as ``w/o reuse new Gaussians'' in \tref{tab:ablation}. 
In the Appendix D.2, we further discuss the enhancement of temporal consistency in dynamic regions.


\begin{figure}[tb!]
    \centering
    \includegraphics[width=0.95\linewidth]{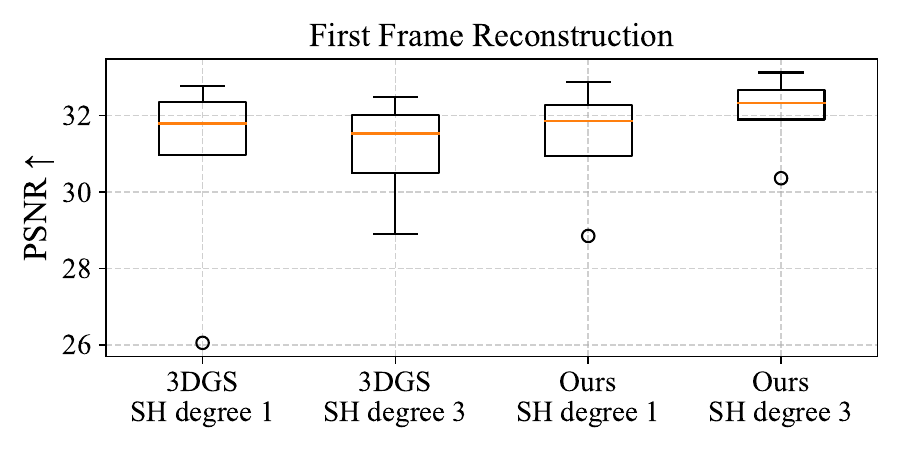}
    \caption{\textbf{Ablation Study of SH Degree}. Difference in first frame reconstruction quality between 3DGS and ours based on SH degree.}
    \label{fig:abl_SH}
\Description[Figure]{spatiotemporal images}
\end{figure}

\section{Conclusion}
\label{sec:conclusion}

In this paper, we reveal that the previous online reconstruction methods do not produce temporally consistent results.
One of the causes of this problem is the errors in the observations.
We resolve the problem by decomposing the rendered images into the ideal signal and the errors during the optimization. Our method significantly improves temporal consistency and visual quality.

\paragraph{Limitation}
Existing baselines struggle under challenging conditions such as motion blur, sparse viewpoints, severe noise, and fast motion. As these failures arise from problem settings that are orthogonal to our temporal consistency objective, our method does not resolve these failures (\fref{fig:limitation}). Moreover, since our method assumes well-reconstructed scenes, its effectiveness decreases in these scenarios (Appendix D.4-D.5). However, our method will remain effective in improving temporal consistency when integrated with future work that improves upon these baselines.

In addition, a more in-depth analysis of potential causes of temporal inconsistency beyond noise, as well as the impact of various types of noise is needed, which we leave for future work.

\begin{figure}[tb!]
    \centering
    \includegraphics[width=\linewidth]{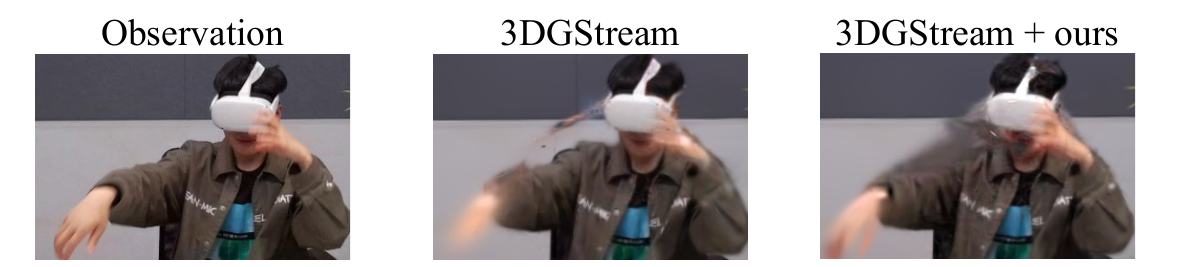}
    \caption{\textbf{Limitation}. Although our method improves the baselines, it does not fix the artifacts due to fast motion.}
    \label{fig:limitation}
\Description[Figure]{Limitation}
\end{figure}

\begin{acks}
This work is supported by the Institute for Information \& Communications Technology Planning \& Evaluation (IITP) grant funded by the Korea government(MSIT) (No. 2017-0-00072, Development of Audio/Video Coding and Light Field Media Fundamental  Technologies  for Ultra Realistic Tera-media)
\end{acks}

    \nocite{schonberger2016colmap,wang2024v}
\else
    
\section{Introduction}
\label{sec:intro}
Reconstructing dynamic scenes from multi-view videos is a crucial problem in computer vision and graphics
to enable freely exploring novel viewpoints and timestamps. This capability has substantial potential for advancing applications in VR, AR, and XR.
Neural Radiance Fields (NeRFs)~\cite{mildenhall2021nerf} have significantly improved the fidelity of 3D reconstruction, and have been adopted to dynamic scenes~\cite{li2022neural}.
However, the relatively slow rendering speed of NeRFs hinders their use in real-time applications.
Recent 3D Gaussian splatting (3DGS)~\cite{kerbl3Dgaussians} has advantages in rapid training and real-time rendering, making it a mainstream approach for dynamic scene reconstruction~\cite{yang2023gs4d,wu20234dgaussians,duan20244drotor}.

However, most dynamic scene reconstruction methods are limited to offline reconstruction, which 
relies on full access to video sequences.
In addition, their results are typically not streamable, which cannot transmit part of the model of the specific moment to the user. Accordingly, they are incompatible with live-streaming applications, where 
only one frame is observable at a moment and the sent frames cannot be updated.
Furthermore, the offline methods face out-of-memory issues with long videos and depend on hyperparameters specific to different lengths.

To reconstruct dynamic scenes from streaming video inputs and enable the streaming of a learned model, recent approaches reconstruct dynamic scenes in online configuration by consecutively optimizing a model which reconstructs a moment~\cite{li2022streaming,sun20243dgstream}.
When the length of the video sequences is undefined, the online methods offer advantages over offline methods due to its unbounded nature~\cite{Wang2023rerf}.
Furthermore, online methods do not rely on length-related hyperparameters.
However, online reconstruction lacks temporal consistency in static regions, as shown in \fref{fig:overview}a. 
We aim to produce temporally consistent and high quality results by finding the problem and solution.

First, we find that the observations bear unnoticeable error rather than capturing the ideal signal as shown in \fref{fig:overview}b. Although we do not know the ideal signal, the colors in different moments at static regions should be the same, while the differences between adjacent frames are non-zero even at static regions. These errors are inevitable due to sensor noises or other reasons. By nature, they vary over time.
We suggest that the online reconstruction produces temporally inconsistent results at static regions because it observes a frame at a moment and overfits to the time-varying errors.
We demonstrate that the errors in the observation harm the temporal consistency through experiments on the synthetic dataset
where we can prepare the observations with and without errors (\sref{sec:method:problem}).

To this end, we propose \textit{observation-restoring online reconstruction} to reconstruct the ideal scene from imperfect observations with errors. 
The key idea is to reconstruct observations combining the Gaussians rendered on the images and residual maps which separately model the errors. We defer the details to \sref{sec:method:decomposing}.
Combined with baselines, our method improves both temporal consistency and the quality of rendered results. The experiments cover various baselines and datasets.
Additionally, our method reduces the number of Gaussians in the reconstruction leading to faster training and rendering, and lower memory footprint. Lastly, the performance measures have lower variance across multiple runs of our method compared to the baselines, indicating stability of our method.


\section{Related Work}
\label{sec:related}
In this section, we briefly review the literature on dynamic scene reconstruction using radiance fields, categorizing approaches into offline (\sref{sec:related:offline}) and online reconstruction (\sref{sec:related:online}). 
Then, we discuss existing works reconstructing the radiance field from inaccurate observations (\sref{sec:related:correction}).

\subsection{Offline Reconstruction of Radiance Fields}
\label{sec:related:offline}
To extend NeRFs~\cite{mildenhall2021nerf} to reconstruct dynamic scenes, several approaches deform rays~\cite{pumarola2020d,park2021nerfies,park2021hypernerf} or expand 3D representations to 4D~\cite{fridovich2023k,cao2023hexplane}, focusing on improving rendering performance, memory efficiency, and training speed.
NeRFPlayer~\cite{song2023nerfplayer} enables the streaming of a learned model by representing each time of the scene with local feature channels.

The emergence of real-time rendering in 3D Gaussian Splatting (3DGS)~\cite{kerbl3Dgaussians} encourages extensive research in primitives-based dynamic scene reconstruction. Similar to Dynamic NeRFs, several works reconstruct dynamic scenes by deforming the Gaussians~\cite{wu20234dgaussians,bae2024ed3dgs,li2024st4dgs} or extending 3D Gaussians to 4D Gaussians~\cite{yang2023gs4d,cho20244dscaffold}. 

Moreover, due to memory constraints, HyperReel~\cite{attal2023hyperreel} and STG~\cite{li2023spacetimegaussians} independently reconstruct each video segment from a single video sequence. This shows visual inconsistencies between the reconstructed results.
SWinGS~\cite{shaw2024swings} segments videos into size-varying segments based on motion intensity, sharing one frame between adjacent segments and training models for each segment. These models are subsequently fine-tuned across total frames to enhance temporal consistency. 
 
However, these offline reconstruction methods cannot process live-streaming input, relying on recorded video input. Moreover, previous temporal consistency enhancement methods require full-frame fine-tuning which is unsuitable for online reconstruction.

\begin{figure*}[tb!]
    \centering
    \includegraphics[width=0.88\linewidth]{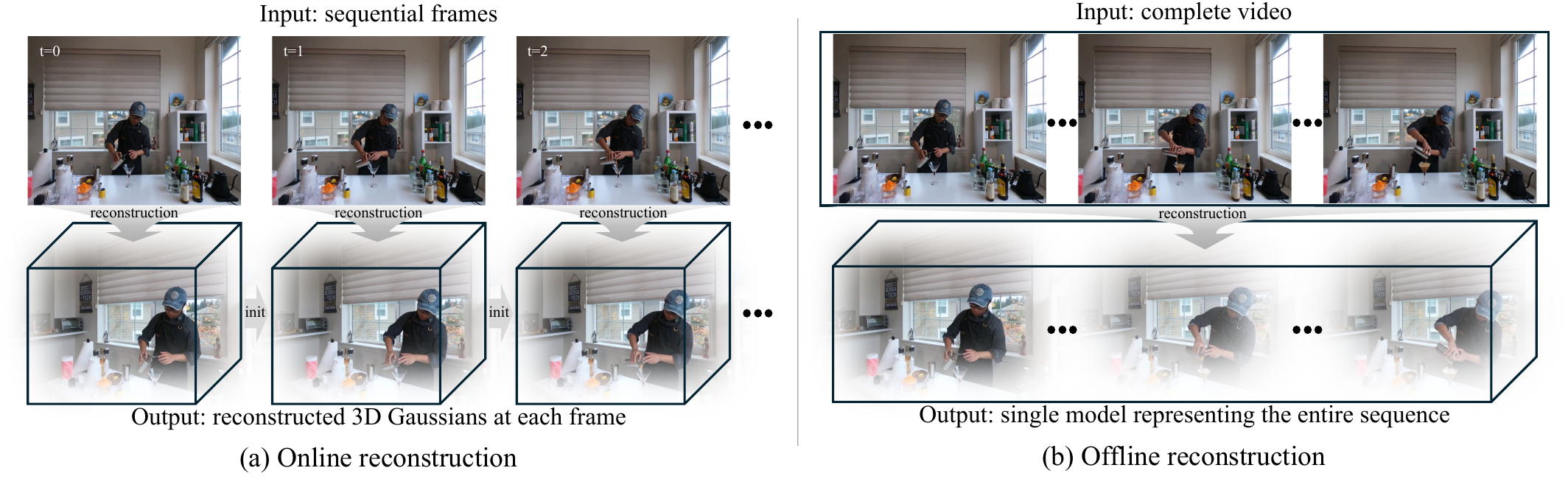}
    \caption{
    \textbf{Comparison of Online and Offline Reconstruction.}
    (a) Online reconstruction takes streaming video as input and sequentially learns a model for each frame.
(b) In contrast, offline reconstruction takes the complete set of video frames as input and produces a single model representing the entire sequence.}
    \label{fig:setting}
\Description[Figure]{Figure of the overall framework}
\end{figure*}

\subsection{Online Reconstruction of Radiance Fields}
\label{sec:related:online}
Online reconstruction methods sequentially optimize the scene at a given time, enabling live-streaming input processing and learned model streaming.
The simplest way for online reconstruction is to train a model from each frame of live-streaming input. However, as most video contents contain significant redundancy across frames, reconstructing each frame independently is inefficient. 
Therefore, StreamRF~\cite{li2022streaming} and ReRF~\cite{Wang2023rerf} consecutively learn the differences from previous frame for efficiency. 

Recent research on online reconstruction predominantly uses 3D Gaussian Splatting (3DGS) as a representation for real-time rendering. These methods optimize the difference between consecutive frame models~\cite{luiten2023dynamic}. To enhance training efficiency, 3DGStream~\cite{sun20243dgstream} employs iNGP~\cite{mueller2022instant}, an implicit network for optimizing the residual instead of directly optimizing Gaussian parameters. HiCoM~\cite{gao2024hicom} introduces an explicit hierarchical motion field to reduce model capacity and utilizes a reference frame prediction to accelerate training. 

However, these studies focus on the efficiency and visual quality of each frame, overlooking the temporal consistency of their results. We aim to enhance the temporal consistency of online reconstruction methods.

\subsection{Optimizing Radiance Field from Imperfect Data}
\label{sec:related:correction}
NeRF$--$~\cite{wang2021nerfmm}, BARF~\cite{lin2021barf}, and SCNeRF~\cite{SCNeRF2021} jointly optimize NeRF and camera parameters, reducing the need for known camera parameters in static scene reconstruction. NoPe-NeRF~\cite{bian2023nopenerf}, CF-3DGS~\cite{fu2024cf3dgs} leverage depth priors for more accurate pose estimation. RoDyNeRF~\cite{liu2023robust} extends this approach by jointly optimizing dynamic NeRF and camera parameters to correct inaccurate camera poses. 
SyncNeRF~\cite{Kim2024Sync} jointly optimizes dynamic radiance fields and time offsets to reconstruct high-quality 4D models from unsynchronized videos.

DeblurNeRF~\cite{li2022deblurnerf}, Deblur3DGS~\cite{lee2024deblurring}, Robust3DGaussians~\cite{darmon2024robust}, and BAD-Gaussians~\cite{zhao2024badgaussians} learn blur formation to obtain sharp reconstruction from blurry inputs. Deblur4DGS~\cite{wu2024deblur4dgs} extends this concept to learning blur formation for sharp dynamic scene reconstruction from blurry monocular videos.
DehazeNeRF~\cite{chen2023dehazenerf}, SeaSplat~\cite{yang2024seasplat}, and DehazeGS~\cite{yu2025dehazegs} reconstruct a clean scene from hazy inputs.
NeRF-W~\cite{martin2021nerf}, WildGaussians~\cite{kulhanek2024wildgaussians}, and NeRF-On-the-go~\cite{Ren2024NeRF} reconstruct 3D scenes from internet images by learning to ignore distractors, such as moving objects.

Sharing a key insight with our method, RawNeRF~\cite{mildenhall2021rawnerf} begins with the premise that all real images contain noise.
RawNeRF demonstrates that NeRF can remove zero mean noise in the observation by averaging information using multi-view correspondence in high dynamic range (HDR) space. However, converting an HDR image where per-pixel noise follows zero-mean Gaussian distribution into a low dynamic range (LDR) image through nonlinear mapping alters the noise distribution, corrupting the noise distribution to a nonzero mean distribution. As a result, noise removal becomes less effective by averaging information in LDR images.

Several methods have attempted to reconstruct from inaccurate inputs, but they have generally focused on modeling errors under specific settings. 
In contrast, we address errors that occur naturally in real-world recording scenarios, where the error factors are diverse and ambiguous, making it challenging to model each factor separately.


\section{Observation-Restoring Online Reconstruction}
\label{sec:method}
In this section, we first provide a preliminary of the online reconstruction of dynamic 3D Gaussian Splatting (\sref{sec:method:prelim}). Next, we introduce our problem setting (\sref{sec:method:problem}). Finally, we propose its solution (\sref{sec:method}).
\algref{alg:overview} in the Appendix summarizes our method.

\subsection{Preliminary: Online Dynamic 3D Gaussian Splatting}
\label{sec:method:prelim}

Offline reconstruction aims to optimize a single model representing the entire sequence from the entire duration of multi-view videos. 
In contrast, online reconstruction aims to reconstruct a model of a \textit{moment} and update the model to sequentially reconstruct consecutive frames, for given sequentially captured multi-view images. This configuration is inevitable for live streaming free-viewpoint videos.
Noting that the offline approaches assume a fixed duration for a model, the online configuration is advantageous for training long and arbitrary-length videos because the length of reconstruction increases while capturing a scene. \fref{fig:setting} conceptually compares online and offline reconstruction.

We provide a formal configuration.
Let $G_t$ be a set of Gaussians at time $t$, defined as $G_t = (\boldsymbol{\mu}_t,\mathbf{q}_t,\mathbf{s}_t,\boldsymbol{\sigma}_t,\mathbf{Y}_t)$ where
$\boldsymbol{\mu}, \mathbf{q}, \mathbf{s}, \boldsymbol{\sigma}$, and $\mathbf{Y}$ denotes mean, rotation, scale, opacity, and spherical harmonics (SH) coefficients of the Gaussians, respectively\footnote{For brevity, we omit the indices of each Gaussian and its parameters.}.
Given a set of cameras $\mathbb{V}$, let $I_t^v$ be the observation of a scene from a camera $v \in \mathbb{V}$ at time $t \in \mathbb{Z}_{\geq 0}$\footnote{$t$ is not bounded above in contrast to offline configuration.}. 

The attributes of the Gaussians $G_0$ are initialized from Structure-from-Motion (SfM) points.
These attributes are optimized by rendering $G_0$ across multiple cameras $v$, minimizing the loss between the rendered image $\hat{I}_0^v$ and their corresponding observation $I_0^v$. 
The number of Gaussians is adaptively adjusted by densification and pruning during optimization. Gaussians are split and cloned if their pixel-space gradient\footnote{Gradient of the loss with respect to the coordinates projected on the image planes} exceeds the predefined threshold, and pruned if their opacity is low. The Gaussians $G_{t+1}$ at the next frame are obtained by deforming the attributes of the optimized $G_t$ from the previous frame.

By design, deforming the Gaussians from previous frames cannot reconstruct new objects.
Therefore, 3DGStream~\cite{sun20243dgstream} introduces new Gaussians $G_t^{\text{new}}$ to represent novel objects.
However, 3DGStream uses these new Gaussians for a single frame and discard them, leading to temporally inconsistent results in dynamic regions (\aref{sup:more:new}).
To solve this, similar to previous works~\cite{gao2024hicom,girish2024queen}, we propagate both new and deformed Gaussians to subsequent frames to enhance temporal consistency.

\begin{figure}[tb!]
    \centering
    \includegraphics[width=\linewidth]{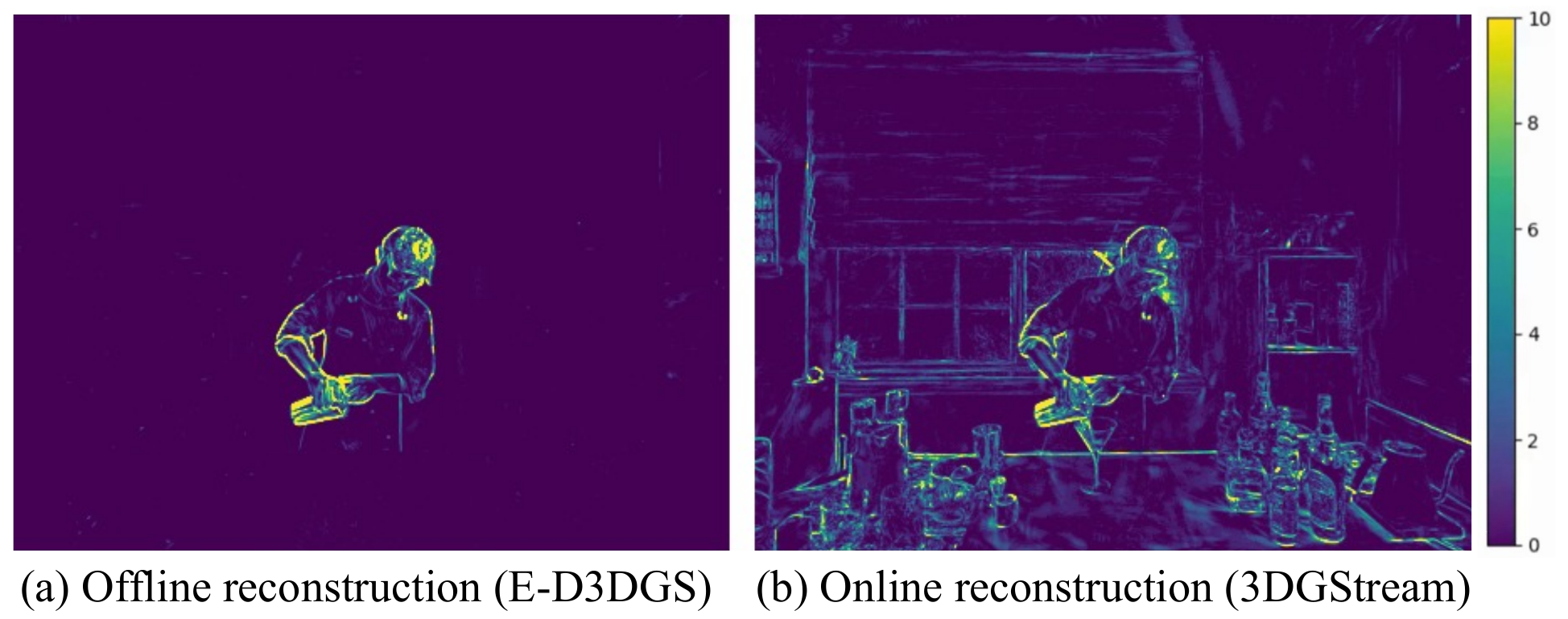}
    \caption{\textbf{Differences between Consecutive Frames.} 
    (a) While the offline reconstruction method maintains temporal consistency in static regions, 
    (b) the online reconstruction method lacks temporal consistency.}
    \label{fig:offline_vs_online}
\Description[Figure]{Different results between offline reconstruction and online reconstruction methods.}
\end{figure}

\begin{figure}[tb!]
    \centering
    \includegraphics[width=\linewidth]{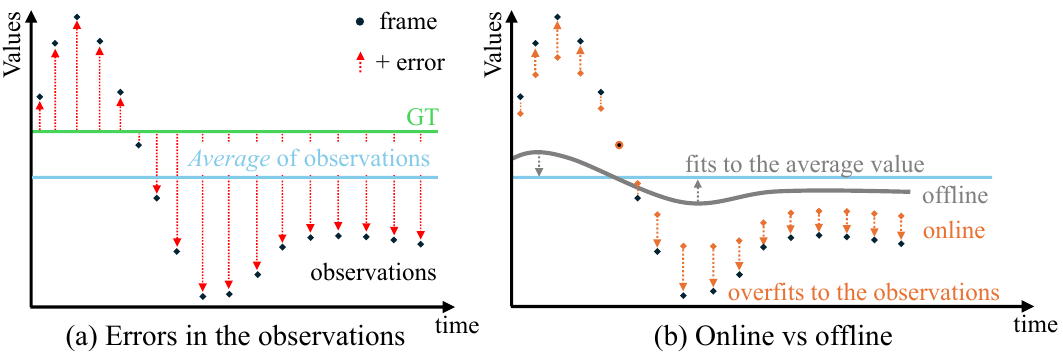}
    \caption{\textbf{Conceptual Figure of the Problem Statement} (a) Each observation value contains errors added to the ground truth. (b) Due to limited data accessibility, online reconstruction tends to overfit each observation, whereas offline reconstruction converges to the global average over time.}
    \label{fig:concept}
\Description[Figure]{The concept of observation errors and the differences between offline and online reconstruction methods.}
\end{figure}

\begin{figure}[tb!]
    \centering
    \includegraphics[width=0.95\linewidth]{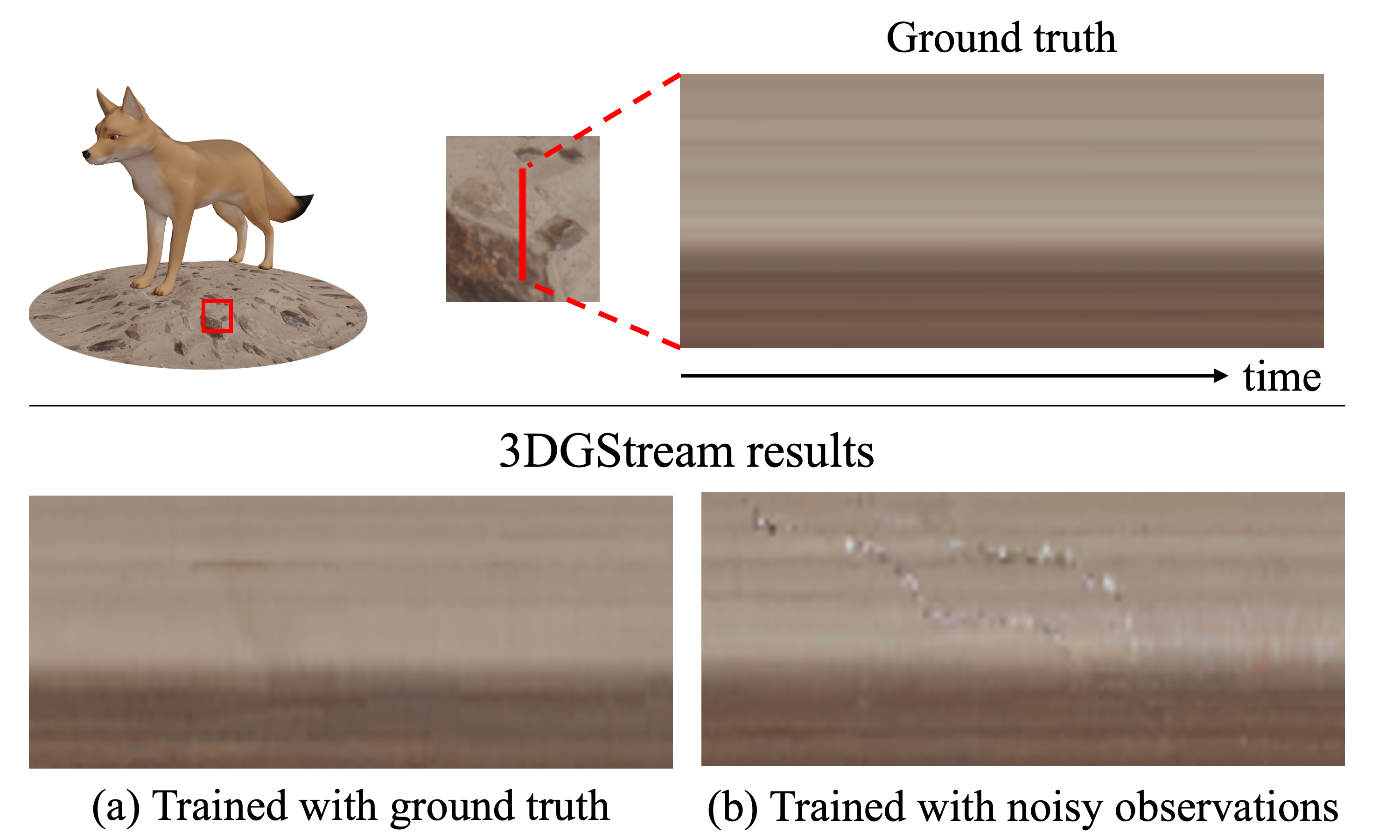}
    \caption{
    \textbf{Comparison of Spatiotemporal Images under Two Settings.}
(a) Training with ground truth images produces clearer results than (b) training with noisy observations created by adding noise to the ground truth. This demonstrates that noise harms the temporal consistency. \copyright DuDeHTM \href{https://creativecommons.org/licenses/by-nc/4.0}{(CC BY-NC 4.0)}
    }
    \label{fig:effect_of_error}
\Description[Figure]{Effect of \error in online reconstruction.}
\end{figure}

\begin{figure}[tb!]
    \centering
    \includegraphics[width=0.95\linewidth]{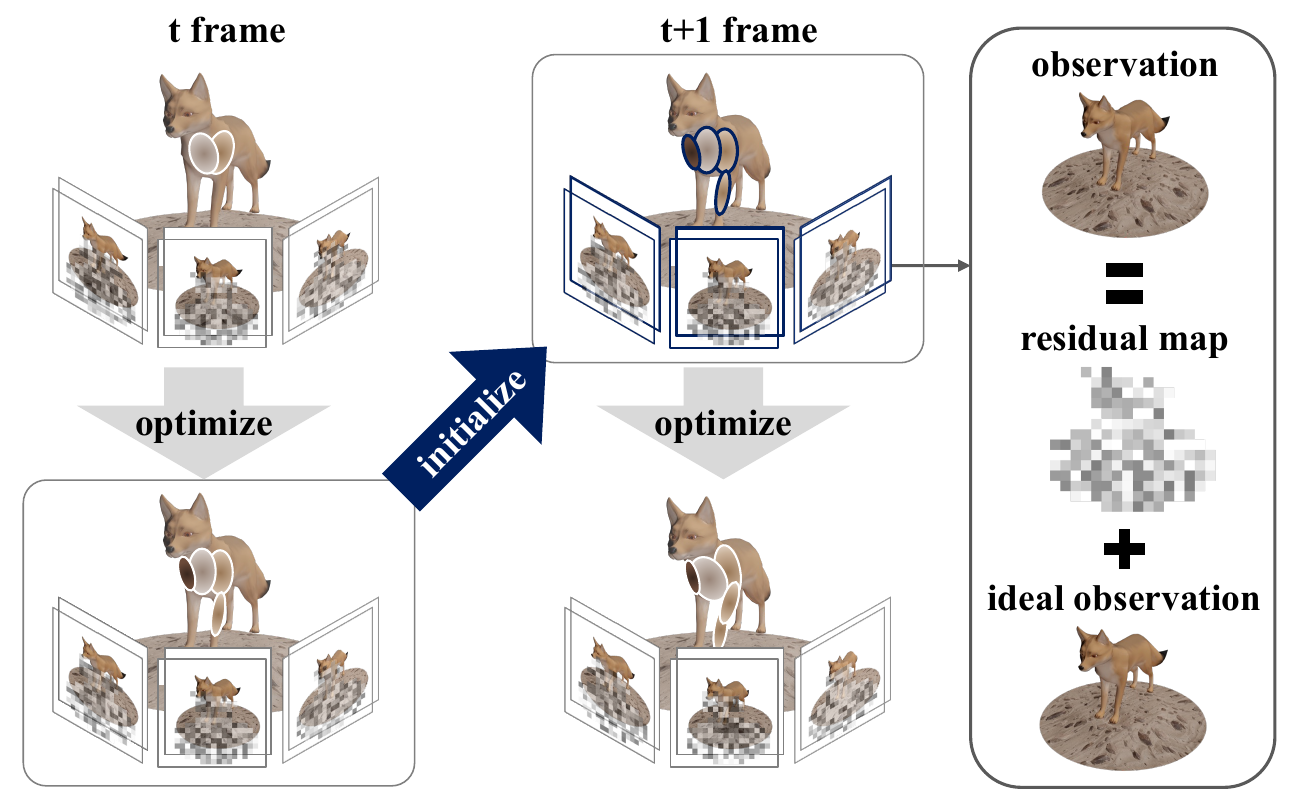}
    \caption{\textbf{Framework.} We jointly optimize 3D Gaussians and residual maps for each camera view. The residual maps model observation errors. For subsequent frames, both Gaussians and residual maps are initialized with values from the previous frame. We model an observation as a combination of the ideal observation and residual map. \copyright DuDeHTM \href{https://creativecommons.org/licenses/by-nc/4.0}{(CC BY-NC 4.0)} 
    }
    \label{fig:framework}
\Description[Figure]{Figure of the overall framework}
\end{figure}

\subsection{Problem Statement}
\label{sec:method:problem}

\begin{figure*}[tb!]
    \centering
    \includegraphics[width=\linewidth]{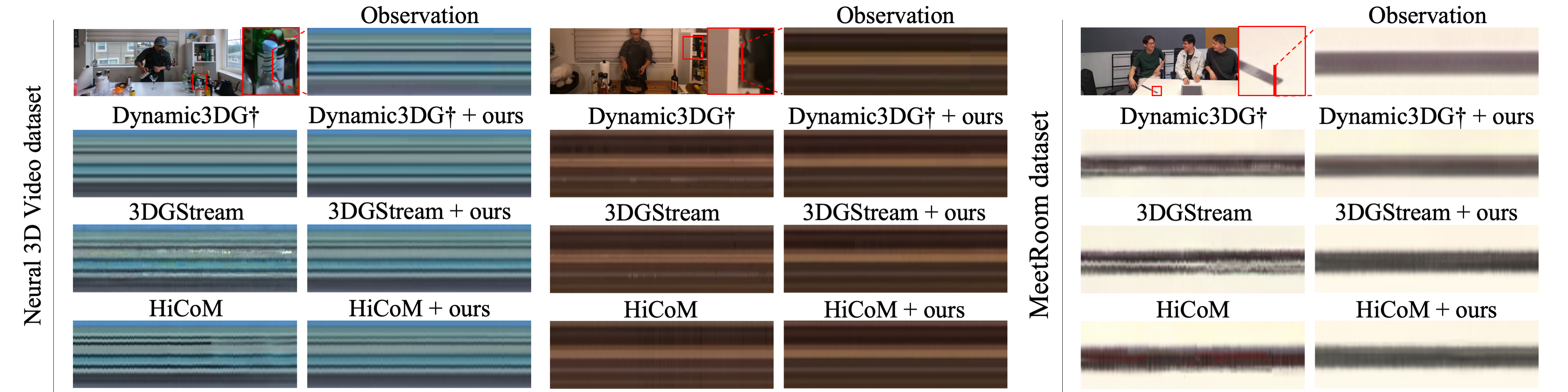}
    \caption{
    \textbf{Qualitative Comparison on Neural 3D Video and MeetRoom Dataset.} Our method produces more temporal consistent spatiotemporal images.}
    \label{fig:qual}
\Description[Figure]{qualitative results of the N3V and MeetRoom dataset}
\end{figure*}

Online reconstruction methods lack temporal consistency whereas offline reconstruction methods do not, as shown in \fref{fig:offline_vs_online}. Intriguingly, the observations lacks temporal consistency between adjacent frames even in static regions as shown in \fref{fig:overview}b. We hypothesize that the observations are corrupted and do not ideally observe the original scene.

Offline reconstruction does not suffer from the above problem because the model converges to the average of all frames because the model with limited capacity cannot fit all tiny differences between the frames as shown in \fref{fig:concept}.
This averaging phenomenon ignores temporally varying errors in the observations, resulting in relatively consistent values for static regions.
In contrast, online reconstruction converges to render a specific frame because it can access only the single target frame.
As a result, the reconstructed result overfits the temporally varying errors in its corresponding observations, leading to temporally inconsistent reconstruction even in static regions.

We support the problem statement with a toy experiment. We reconstruct a synthetic dynamic 3D asset for given observations prepared by rendering it on fixed viewpoints, with and without noise\footnote{We add Gaussian and Poisson noise.}. While the results from ground truth observations are mostly clear (\fref{fig:effect_of_error}a), the results from noisy observation exhibit flickering artifacts (\fref{fig:effect_of_error}b).

We formulate the observations $\tilde{I}_t^v$ capturing the scene as a combination of ideal observations $I_t^v$ and errors $M_t^v$, and propose to restore the ideal scene with temporal consistency by disregarding the errors.

\subsection{Decomposing Errors in the Observations}
\label{sec:method:decomposing}
\paragraph{Learnable residual map}
\label{sec:method:residual}
To address observations containing errors that vary across camera views $v$ and frames $t$, we separately model these errors with additional residual maps $\hat{M}_t^v \in \mathbb{R}^{3 \times H \times W}$, which are optimized as learnable parameters during training:
\begin{equation}
\label{eq:separate}
    \tilde{I_t}=I_t+M_t=\hat{I}_t+\hat{M}_t,    
\end{equation}
where $\hat{I}_t$ and $\hat{M}_t$ are rendered image and estimated residual, respectively, and we omit $v$ for brevity. $H$ and $W$ denote the height and width of the image $I_t$.
During optimization, we minimize the difference between $\tilde{I_t}$ and $\hat{I}_t+\hat{M}_t$ by jointly updating $G_t$ and $\hat{M}_t$. An optimization step becomes:
\begin{equation}
    G_t, \hat{M_t} \gets \text{Adam}(\nabla L_\text{total}(\tilde{I}_t,\hat{I}_t+\hat{M_t})),
\label{eq:backward}
\end{equation}
where $L_\text{total}$ will be described later.

Separately modeling the residual maps restores the ideal observations as follows.
In \eref{eq:separate}, we expect the rendered images and the estimated residual maps to match the corrupted observation together. 
Fitting multi-view inconsistent high-frequency noise with a residual maps are easier than optimizing one set of Gaussians. Moreover, learning multi-view and temporal inconsistent noise with a residual map is easier than fine-tuning a well-trained Gaussians from the previous frame.
As a result, the optimized Gaussians reconstruct the true scene without corruption. I.e., the rendered images become the ideal observation without errors.

\paragraph{First Frame Reconstruction}
\label{sec:method:init}
To reconstruct the first frame, we initialize Gaussians from SfM points~\cite{schonberger2016colmap} and residual maps to zero.
We then jointly optimize them same as 3DGS~\cite{kerbl3Dgaussians}. To prevent the residual maps from dominating the reconstruction, they are frozen at zero until the Gaussians start densification.

In addition, we apply L1 regularization on opacity~\cite{kheradmand2024gsmcmc,bae2024ed3dgs} as resetting opacity in 3DGS~\cite{kerbl3Dgaussians} disrupts stable optimization of the residual maps: 
$L_\text{opa} = \sum_i{ \|\sigma_t^i\|_1 }$,
where $\sigma_t^i$ denotes opacity of the $i$-th Gaussian at time $t$.
We also employ L1 regularization on the residual map to prevent overfitting to view-conditioned color:
$L_\text{res} = \|\hat{M}_t^v\|_1.$

The total loss for the first frame is:
\begin{equation}
    L_\text{total} = (1 - \lambda)L_1 + \lambda L_\text{D-SSIM} + \lambda_\text{opa} L_\text{opa} + \lambda_\text{res} L_\text{res}.
\label{eq:loss:first}
\end{equation}

\paragraph{Sequential Frame Reconstruction}
\label{sec:method:per-frame}
We consecutively propagate new Gaussians across subsequent frames: $G_{t+1} \gets (G_t,G_t^\text{new})$ because using the new Gaussians in only a single frame and discarding them harms the temporal consistency.
We do not use $L_\text{opa}$ in sequential frame reconstruction to ensure fair comparison with baselines.
Finally, the total loss for the remaining frames is:    
\begin{equation}
    L_\text{total} = (1 - \lambda)L_1 + \lambda L_\text{D-SSIM} + \lambda_\text{res} L_\text{res}.
\label{eq:loss:seq}
\end{equation}

\begin{table*}[tbh!]
    \centering
    \caption{\textbf{Average Performance on Neural 3D Video and MeetRoom Dataset}. Computational cost was measured on \texttt{flame\_salmon} with RTX A5000.}
    \resizebox{0.9\linewidth}{!}{
    \begin{tabular}{l|ccc|ccc|cc}
    \toprule
    \multirow{2}{*}{Model}  &   \multicolumn{3}{c|}{Neural 3D Video} &   \multicolumn{3}{c|}{MeetRoom} &   \multicolumn{2}{c}{Computational Cost}  \\               
    & PSNR $\uparrow$ & SSIM $\uparrow$ & mTV$_{\times100}$ $\downarrow$ & PSNR $\uparrow$ & SSIM $\uparrow$ & mTV$_{\times100}$ $\downarrow$ & \# Gaussians $\downarrow$ & Training time $\downarrow$ \\
    \midrule
    Dynamic3DG$\dagger$ & 32.48 & 0.960 & 0.153 & 30.79 & 0.952 & 0.175 & 315K & 136. sec/frame         \\ 
    Dynamic3DG$\dagger$ + ours  & \textbf{33.03} & \textbf{0.961} & \textbf{0.053} & \textbf{31.29} & \textbf{0.953} & \textbf{0.126} & \textbf{246K} & \textbf{130.} sec/frame         \\ \midrule
    3DGStream           & 32.58 & 0.960 & 0.178 & 31.78 & 0.955 & 0.049 & 315K & 13.3 sec/frame         \\
    3DGStream + ours    & \textbf{33.13} & \textbf{0.961} & \textbf{0.103} & \textbf{32.29} & \textbf{0.957} & \textbf{0.024} & \textbf{268K} & \textbf{13.1} sec/frame        \\ \midrule
    HiCoM               & 32.09 & 0.956 & 0.164 & 29.42 & 0.936 & 0.127 & 307K & 11.0 sec/frame         \\
    HiCoM + ours        & \textbf{32.46} & \textbf{0.957} & \textbf{0.118} & \textbf{29.48} & \textbf{0.938} & \textbf{0.045} & \textbf{218K} & \textbf{10.2} sec/frame        \\ 
    \bottomrule
    \end{tabular}
    }
    \label{tab:quant}
\end{table*}

\section{Experiment}
\label{sec:exp}
In this section, we first provide the reasons for choosing the dataset, implementation, and the evaluation metrics (\sref{exp:criteria}). 
Next, we demonstrate our improvements in both temporal consistency and visual quality upon the baselines (\sref{sec:exp:comparison}).
We then show our proposed method effectively restores the observations (\sref{sec:exp:restoration}). 
Finally, we conduct an ablation study to assess the performance of our method (\sref{sec:exp:ablation}).


\subsection{Datasets, Implementation, and Metrics}
\label{exp:criteria}
\subsubsection{Datasets}
\label{exp:criteria:data}
Following the prior works~\cite{luiten2023dynamic,sun20243dgstream}, we use undistorted images and SfM points in our experiments.  Additional experiments on the original images and MVS points~\cite{wu20234dgaussians,gao2024hicom} are detailed in \aref{sup:offline}. We evaluated the baselines on these datasets.
\paragraph{Neural 3D Video~\cite{li2022neural}} The dataset contains six scenes captured by 17 to 21 multi-view cameras over 300 frames.
Following the prior works~\cite{wang2022mixed,fridovich2023k}, we downsample the video by a factor of two and remove the unsynchronized video in \texttt{coffee\_martini} scene.

\paragraph{MeetRoom~\cite{li2022streaming}} The dataset contians three scenes captured by 14 multi-view cameras over 300 frames.

\paragraph{Sync-NeRF~\cite{Kim2024Sync}} The dataset is rendered from dynamic 3D asset with static regions, providing a training dataset without errors (ground truth dataset). 
For our experiments, we use the modified \texttt{fox}\footnote{\copyright DuDeHTM \href{https://creativecommons.org/licenses/by-nc/4.0}{(CC BY-NC 4.0)} } scene (see \aref{sup:detail:data} for details). We train with noisy observations\footnotemark[1] and evaluate against the ground truth.


\begin{figure}[tb!]
    \centering
    \includegraphics[width=\linewidth]{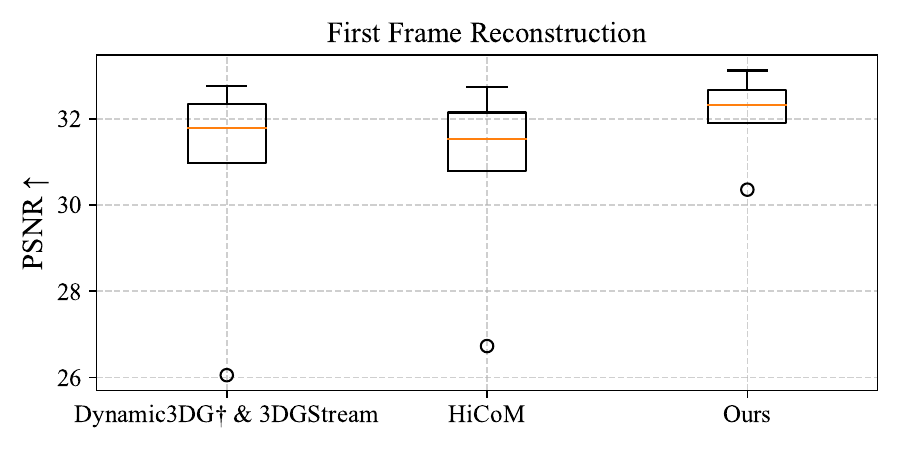}
    \caption{\textbf{First Frame Reconstruction}. Box plots comparing first frame reconstruction quality across 10 runs for each scene in the Neural 3D Video dataset.}
    \label{fig:initial_frame}
\Description[Figure]{Initial frame reconstruction}
\end{figure}

\subsubsection{Implementation}
\label{exp:criteria:imple}
We apply our method to the state-of-the-art online reconstruction methods (Dynamic3DG~\cite{luiten2023dynamic}, 3DGStream~\cite{sun20243dgstream}, and HiCoM~\cite{gao2024hicom}) and compare the performance with the original implementations. StreamRF \cite{li2022streaming} is excluded because it is hardly reproducible.

We follow the official implementations except for Dynamic3DG. Due to the poor results of Dynamic3DG on the datasets with forward-facing configuration, we modified it to improve performance, naming it Dynamic3DG$\dagger$ (see \aref{sup:detail:imple} for details). To ensure a fair comparison, we apply our method to HiCoM using the same pruning strategy proposed in it\footnote{HiCoM maintains compactness by pruning the K lowest opacity Gaussians from the combined set of deformed and new Gaussians, where K equals the number of new Gaussians. However, this pruning strategy degrades visual quality.}.

As 3DGS~\cite{kerbl3Dgaussians} has randomness in optimization\footnote{Please refer to https://github.com/graphdeco-inria/gaussian-splatting/issues/89}, we select the 3D Gaussians with the highest PSNR from 10 runs as the initialization for reconstructing the sequential frame. This protocol is confirmed by correspondence with the authors of 3DGStream.
We use the same 3D Gaussians at the first frame for both Dynamic3DG$\dagger$ and 3DGStream\footnote{Note that once the 3D Gaussians at the first frame are fixed, sequential frames bear minimal variance, indicating that the randomness in online reconstruction primarily derives from the first frame.}.

In our experiments, we do not prune Gaussians with large radii in pixel-space because it degrades performance.
We set the SH degree to 3 when applying our method (See \sref{sec:exp:ablation}).


\subsubsection{Metrics}
\label{exp:criteria:metric}
We report the quality of rendered images using
Peak Signal-to-Noise Ratio (PSNR), Structural Similarity Index Measure (SSIM), and masked Total Variation (mTV).
PSNR quantifies pixel color error between the rendered video and the test dataset. 
We measure SSIM to account for the perceived similarity of the rendered image. Additionally, 
we measure mTV to measure temporal consistency in static regions indicated by predefined masks. \aref{sup:detail:data} provides details about the masks used in the experiments. Higher PSNR and SSIM values and lower mTV values indicate better visual quality. For readability, we multiply mTV by 100 and refer to it as mTV$_{\times100}$. 
 

\subsection{Comparison on Real-world Dataset}
\label{sec:exp:comparison}

\subsubsection{Visual Quality}
\label{sec:exp:comparisions:init}

\paragraph{The First Frame} Our method improves PSNR at the first frame in both median and highest values as shown in \fref{fig:initial_frame}. The box plot is drawn with 10 runs per scene in the Neural 3D Video dataset. In addition, our method achieve lower variance, implying higher stability than the baselines.

\paragraph{All Frames}
Consecutively applying our method to all frames improves both PSNR and SSIM, achieving superior visual quality as shown in \tref{tab:quant}. 

\begin{figure}[tb!]
    \centering
    \includegraphics[width=\linewidth]{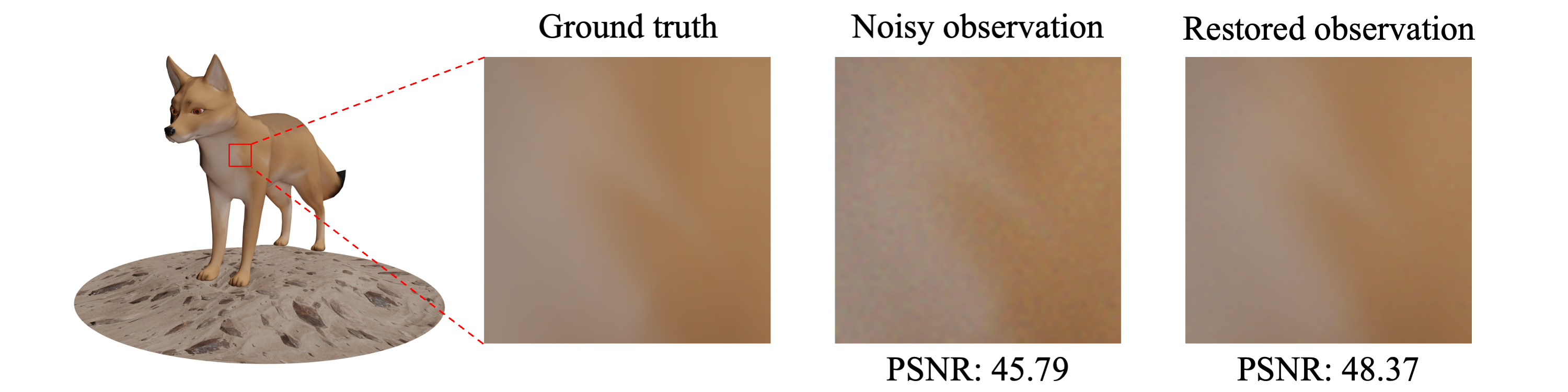}
    \caption{\textbf{Observation Restoration}. Our learned residual maps successfully restore the noisy observations. \copyright DuDeHTM \href{https://creativecommons.org/licenses/by-nc/4.0}{(CC BY-NC 4.0)}
    }
    \label{fig:restoration}
\Description[Figure]{Restoration.}
\end{figure}
\begin{table}[tb!]
    \centering
    \caption{\textbf{Average Performance on Sync-NeRF Dataset}.}
    \resizebox{\linewidth}{!}{
    \begin{tabular}{l|ccc}
        \toprule
        Model & PSNR $\uparrow$ & SSIM $\uparrow$ & mTV$_{\times100}$ $\downarrow$  \\
        \midrule
         3DGStream & 38.79 & 0.982 & 0.483 \\
         3DGStream on denoised observations & 38.63 & 0.979 & 0.417 \\
         3DGStream + ours & \textbf{39.16} & \textbf{0.987} & \textbf{0.318} \\
        \bottomrule
    \end{tabular}
    }
    \label{tab:synthetic}
\end{table}


\subsubsection{Temporal Consistency}
\fref{fig:qual} presents spatiotemporal images of static areas. While baseline methods show noisy results, combining our method produces clear outputs, indicating improved temporal consistency. 
In \tref{tab:quant}, applying our method reduces mTV, demonstrating enhanced temporal consistency.


\subsubsection{Computational Cost}
\label{sec:exp:comparisions:cost}
\tref{tab:quant} shows the computational costs.
Our approach reduces the number of Gaussians for each baselines. 
Although we optimize additional parameters, the reduced number of Gaussians decreases the training time.
Notably, the residual maps are only used during training and are not stored.


\subsection{Comparison on Synthetic Dataset}
\label{sec:exp:restoration}
\subsubsection{Restoration}
To demonstrate the ability of our learned residual map to restore the observations, we conduct experiments on Sync-NeRF dataset. 
We restore the train view observation to its ground truth by subtracting the residual map from the corresponding observation.
This restoration increases PSNR (+2.58) with clear visual quality, as shown in \fref{fig:restoration}.

\subsubsection{Comparison to Preprocessing}
We additionally compare our method against combination of 3DGStream and a denoising model \cite{liang2022vrt}, where 3DGStream reconstructs the scene from the denoised observations.

As shown in \tref{tab:synthetic}, our method outperforms competitors in both visual quality (PSNR) and temporal consistency (mTV). Using a denoising model also enhances temporal consistency, which supports our problem settings. Nonetheless, its effectiveness remains lower than ours\footnote{Moreover, \citet{liang2022vrt} degrade visual quality through over-smoothing, and the preprocessing time (196 seconds per frame) significantly exceeds the sequential frame reconstruction time (6 seconds per frame).}.

\begin{table}[tb!]
\centering
\caption{\textbf{Quantitative Comparison of Ablation Study.}}
\resizebox{0.85\linewidth}{!}{
\begin{tabular}{l|ccc}
\toprule
    Method & PSNR $\uparrow$ & SSIM $\uparrow$ & mTV$_{\times100}$ $\downarrow$ \\
\midrule
3DGStream + ours & \textbf{33.13} & 0.961 & \textbf{0.103} \\ \midrule
w/o residual maps & 32.65 & 0.960 & 0.161 \\
w/o SH degree 3 & 33.08 & 0.961 & 0.129 \\
w/o reuse new Gaussians & 33.06 & 0.960 & 0.118 \\
\bottomrule
\end{tabular}
}
\label{tab:ablation}
\end{table}
\begin{figure}[tb!]
    \centering
    \includegraphics[width=0.95\linewidth]{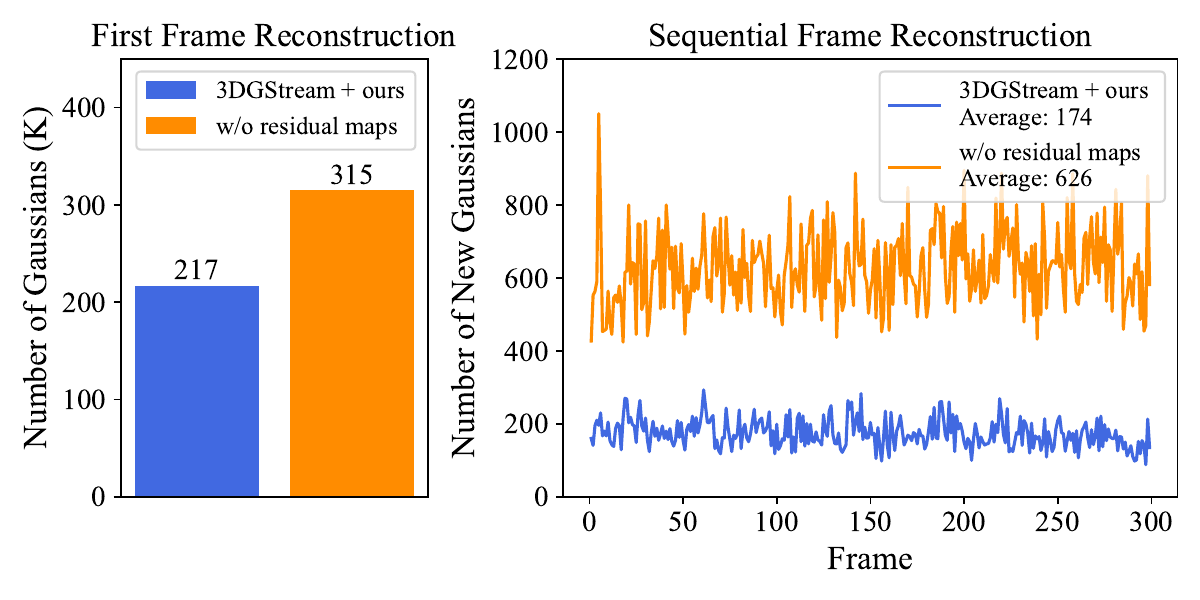}
    \caption{\textbf{Ablation Study of Residual Maps}. Our proposed residual maps reduces the number of Gaussians in both first frame reconstruction and subsequent new Gaussian additions, measured on the \texttt{flame\_salmon} scene.}
    \label{fig:abl_new_gaussian}
\Description[Figure]{spatiotemporal images}
\end{figure}


\subsection{Ablation Study}
\label{sec:exp:ablation}
\subsubsection{Learnable Residual Maps}
To validate the effect of the residual map, we compared ``3DGStream + ours'' to a version without the residual map, labeling ``w/o residual maps''.
As shown in Table 2, using residual maps improves both temporal consistency (reduced mTV) and visual quality (increased PSNR).
Our method reduces unnecessary Gaussians by separating observation errors that inflate the pixel-space gradient.
Leveraging our residual maps decreases the number of Gaussians, reducing $G_0$ to 0.69$\times$ and $G_t^{\text{new}}$ to 0.28$\times$, as shown in \fref{fig:abl_new_gaussian}.

\subsubsection{SH Degree}
We analyze the variation of reconstruction performance due to different SH degrees.
The higher the SH degrees, the higher the model capacity.
In the first frame reconstruction, 3DGS~\cite{kerbl3Dgaussians} with higher model capacity tends to overfit observation errors, performing better at SH degree 1, as shown in \fref{fig:abl_SH}. 
In contrast, our residual maps fit these errors instead of SH, allowing Gaussians to effectively use their higher model capacity without overfitting the errors. Moreover, higher 3D Gaussian capacity prevents residual maps from overfitting view-dependent color, thus achieving superior performance at SH degree 3. 
Consequently, in sequential frame reconstruction, setting the SH degree to 3 enhances visual quality and temporal consistency compared to lower SH degree 1, as demonstrated in \tref{tab:ablation} ``w/o SH degree 3''.

\subsubsection{New Gaussians}
Propagating new and deformed Gaussians to the next frame enhances visual quality and temporal consistency in static regions compared to using new Gaussians only in specific frames labeled as ``w/o reuse new Gaussians'' in \tref{tab:ablation}. 
In the \aref{sup:more:new}, we further discuss the enhancement of temporal consistency in dynamic regions.


\begin{figure}[tb!]
    \centering
    \includegraphics[width=0.95\linewidth]{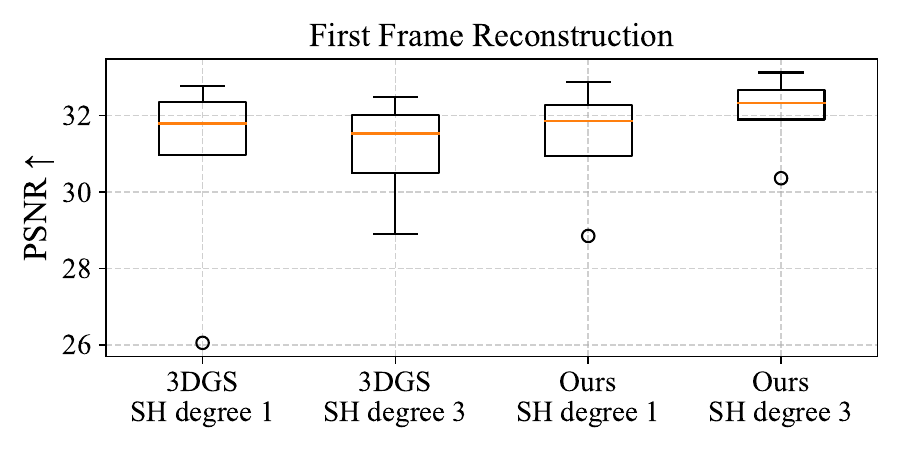}
    \caption{\textbf{Ablation Study of SH Degree}. Difference in first frame reconstruction quality between 3DGS and ours based on SH degree.}
    \label{fig:abl_SH}
\Description[Figure]{spatiotemporal images}
\end{figure}

\section{Conclusion}
\label{sec:conclusion}

In this paper, we reveal that the previous online reconstruction methods do not produce temporally consistent results.
One of the causes of this problem is the errors in the observations.
We resolve the problem by decomposing the rendered images into the ideal signal and the errors during the optimization. Our method significantly improves temporal consistency and visual quality.

\paragraph{Limitation}
Existing baselines struggle under challenging conditions such as motion blur, sparse viewpoints, severe noise, and fast motion. As these failures arise from problem settings that are orthogonal to our temporal consistency objective, our method does not resolve these failures (\fref{fig:limitation}). Moreover, since our method assumes well-reconstructed scenes, its effectiveness decreases in these scenarios (\aref{sup:more:noise}-\ref{sup:more:view}). However, our method will remain effective in improving temporal consistency when integrated with future work that improves upon these baselines.

In addition, a more in-depth analysis of potential causes of temporal inconsistency beyond noise, as well as the impact of various types of noise is needed, which we leave for future work.

\begin{figure}[tb!]
    \centering
    \includegraphics[width=\linewidth]{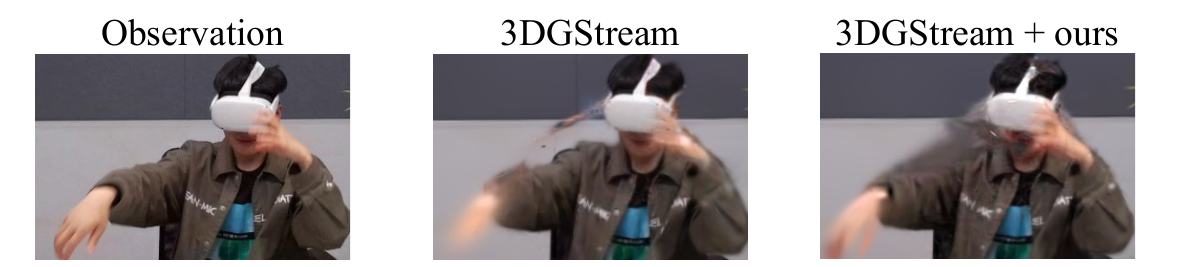}
    \caption{\textbf{Limitation}. Although our method improves the baselines, it does not fix the artifacts due to fast motion.}
    \label{fig:limitation}
\Description[Figure]{Limitation}
\end{figure}

\begin{acks}
This work is supported by the Institute for Information \& Communications Technology Planning \& Evaluation (IITP) grant funded by the Korea government(MSIT) (No. 2017-0-00072, Development of Audio/Video Coding and Light Field Media Fundamental  Technologies  for Ultra Realistic Tera-media)
\end{acks}

\fi
\bibliographystyle{ACM-Reference-Format}
\bibliography{bibliography}

\ifcameraready
\else
    \clearpage
\appendix

\section{Overview}
\label{sup:overview}
\algref{alg:overview}, summarizes our method overview with novel components highlighted in \textcolor{blue}{blue}.

\begin{algorithm}[thb]
\caption{Overview}
\SetKwInOut{Input}{Input}\SetKwInOut{Output}{Output}
\Input{Streaming multi-view images $\tilde{I}_t^v$ at view $v$ and time $t$}
\Output{Optimized Gaussians $G_t$ at time $t$}
$t \gets 0$ \tcp*[r]{video starts}
$G_0 \gets \text{InitAttributes(SfM Points)}$ \tcp*[r]{Intialize attributes}
\textcolor{blue}{\{$M_0^v | v \in V\} \gets \text{InitResidualMap()}$ \tcp*[r]{Zero Init}}
\While{Gaussians $G_0$ not converged}{
    $\hat{I}_0^v \gets$ Rasterize($G_0$, $v$) \tcp*[r]{Rendering}	
    $\bar{I}_0^v \gets \tilde{I}_0^v \textcolor{blue}{- M_0^v} $ \textcolor{blue}{\tcp*[r]{Restore (\fref{fig:framework})}}
    $L \gets L_\text{total}(\hat{I}_0^v, \bar{I}_0^v) $ \textcolor{blue}{\tcp*[r]{Loss (\eref{eq:loss:first})}}
    $G_0 \textcolor{blue}{, M_0^v} \gets$ Adam($\nabla L$) \tcp*[r]{Backprop \& Step (\textcolor{blue}{\eref{eq:backward}})}
    \If{DensificationIteration}{
        Densification() \tcp*[r]{Prune \& Split \& Clone}
    }
}
\While{multi-view video stream is not terminated}{
    $G_{t+1} \gets G_t$ \tcp*[r]{Initialize from previous Gaussians}
    \While{Gaussians $G_t$ not converged}{
        $\hat{I}_{t+1}^v \gets$ Rasterize($G_{t+1}$, $v$) \tcp*[r]{Rendering}	
        $\bar{I}_{t+1}^v \gets \tilde{I}_{t+1}^v \textcolor{blue}{- M_{t+1}^v} $ \textcolor{blue}{\tcp*[r]{Restore (\fref{fig:framework})}}
    $L \gets L_\text{total}(\hat{I}_{t+1}^v, \bar{I}_{t+1}^v) $ \textcolor{blue}{\tcp*[r]{Loss (\eref{eq:loss:seq})}}
        $G_{t+1} \textcolor{blue}{, M_{t+1}^v} \gets$ Adam($\nabla L$) \tcp*[r]{Backprop \& Step (\textcolor{blue}{\eref{eq:backward}})}
    }
    $G_{t+1}^{new} \gets$ SpawnGaussians($G_{t+1}$) \tcp*[r]{Spawn}
    \While{new Gaussians $G_{t+1}^{new}$ not converged}{
        $\hat{I}_{t+1}^v \gets$ Rasterize($(G_{t+1}, G_{t+1}^{new})$, $v$) \tcp*[r]{Rendering}	
        $\bar{I}_{t+1}^v \gets \tilde{I}_{t+1}^v \textcolor{blue}{- M_{t+1}^v} $ \textcolor{blue}{\tcp*[r]{Restore (\fref{fig:framework})}}
    $L \gets L_\text{total}(\hat{I}_{t+1}^v, \bar{I}_{t+1}^v) $ \textcolor{blue}{\tcp*[r]{Loss (\eref{eq:loss:seq})}}
        $G_{t+1}^{new} \textcolor{blue}{, M_t^v} \gets$ Adam($\nabla L$) \tcp*[r]{Backprop \& Step (\textcolor{blue}{\eref{eq:backward}})}
    }
    $t \gets t+1$ \tcp*[r]{next time step}
    \textcolor{blue}{$G_t \gets (G_t,G_t^{new})$ \tcp*[r]{reuse new gaussians}}
}
\label{alg:overview}
\end{algorithm}

\section{Experiments Details}
\label{sup:detail}
\subsection{Implementation Details}
\label{sup:detail:imple}
We implemented our method within the official repositories of each baseline. For the initial frame reconstruction, we trained for 15000 iterations, with densification running up to 7500 iterations for SfM initialization and 5000 iterations for MVS initialization~\cite{schonberger2016colmap}.
During training, we decayed the learning rate for residual maps from 1e-4 to 1e-6 for initial frame reconstruction and from 1e-5 to 1e-7 for sequential frame reconstruction. Both $\lambda_\text{opa}$ and $\lambda_\text{res}$ were set to 0.01. All models were trained with a single RTX A5000.

\subsection{Modified Dynamic3DG}
\label{sup:detail:D3DG}
We modified several aspects of the original implementation of Dynamic3DG~\cite{luiten2023dynamic} that were limiting performance. 
First, instead of using SH degree 0 with per-camera colors to model view-dependent color, we increase the SH degree, as the original approach degraded temporal consistency. 
Second, we removed the velocity-based displacement initialization, which was causing artifacts.
While Dynamic3DG optimizes color differently from the original paper, we do not optimize it to avoid excessive storage from a high SH degree. Instead we introduce a stage to train new Gaussians, which are then propagated to the next frame.

Dynamic3DG$\dagger$ achieves a PSNR of 32.48 on the Neural 3D Video dataset, surpassing the previously reported Dynamic3DG performance of 30.67~\cite{li2023spacetimegaussians}.

\subsection{Details of Datasets}
\label{sup:detail:data}
\subsubsection{Sync-NeRF}
\label{sup:detail:data:syncnerf}
To better distinguish between static and dynamic regions, we modified the scene to eliminate shadows. We then re-rendered the modified assets using the Cycles engine with 512 sample counts to remove rendering noise. We verified the absence of rendering noise by confirming that pixel values in static regions remained constant across all frames.
\subsubsection{Measuring Total Variation in Static Regions}
\label{sup:detail:data:mask}
We manually created masks by defining bounding boxes around dynamic regions to measure total variation in the static regions, as shown in \fref{fig:mask}.

\begin{figure}[tb!]
    \centering
    \includegraphics[width=\linewidth]{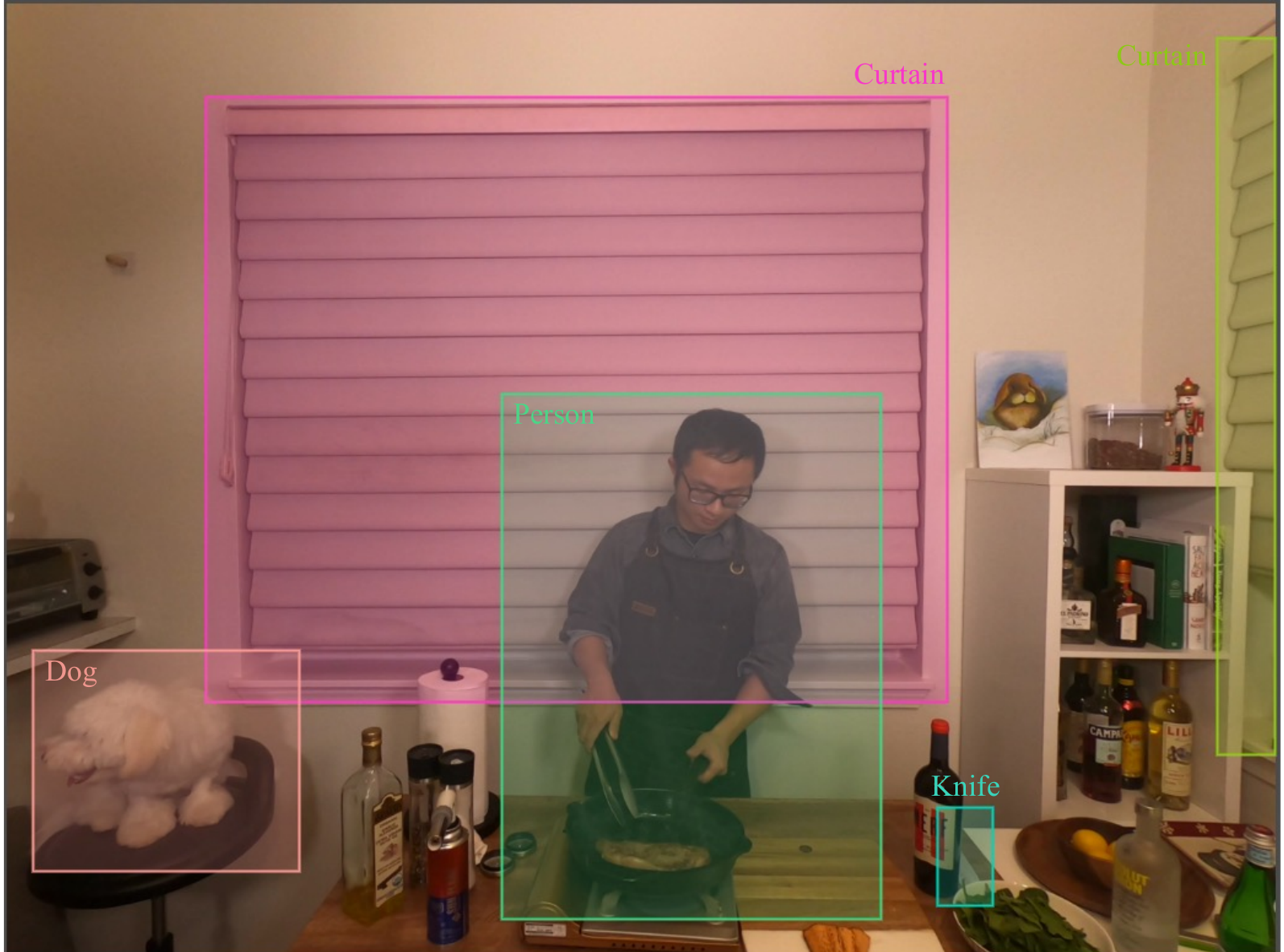}
    \caption{
        \textbf{Bounding Box of the Dynamic Region in the \texttt{sear\_steak} Scene}.
    }
    \label{fig:mask}
\Description[Figure]{Mask}
\end{figure}

\begin{table}[tbh]
    \centering
    \caption{\textbf{Additional Comparison with Offline Reconstruction}. For Online$^*$, we report the average values across 10 runs.}
    \resizebox{\linewidth}{!}{
    \begin{tabular}{c|l|ccc}
    \toprule
    \multirow{2}{*}{Category}   &   \multirow{2}{*}{Model}  &   \multicolumn{3}{c}{Neural 3D Video}  \\ 
                                &                           & PSNR $\uparrow$ & SSIM $\uparrow$ & mTV$_{\times100}$ $\downarrow$ \\
    \midrule
    \multirow{6}{*}{Offline}    &   MixVoxels-L             & 30.81 & 0.933 & \textbf{0.000}         \\
                                &   K-Planes-H              & 31.22 & 0.947 & 0.055         \\ \cmidrule{2-5}
                                &   4DGS                    & 32.14 & 0.947 & 0.036         \\
                                &   4DGaussians             & 31.84 & 0.946 & 0.024         \\
                                &   STG                     & 31.96 & 0.948 & 0.014         \\
                                &   E-D3DGS                 & 32.10 & 0.951 & 0.026         \\
    \midrule
    \multirow{6}{*}{Online$^*$} &   Dynamic3DG$\dagger$     & 31.88 & 0.952 & 0.174        \\
                                &   Dynamic3DG$\dagger$ + ours      & 32.61 & 0.955 & 0.077        \\ \cmidrule{2-5}
                                &   3DGStream               & 31.65 & 0.949 & 0.285        \\
                                &   3DGStream + ours        & \textbf{32.67} & 0.955 & 0.148         \\ \cmidrule{2-5}
                                &   HiCoM                   & 31.61 & 0.948 & 0.193         \\
                                &   HiCoM + ours            & 31.75 & 0.948 & 0.129         \\
    \bottomrule
    \end{tabular}
    }
    \label{tab:n3v_w_offline}
\end{table}

\begin{table}[tbh]
    \centering
    \caption{\textbf{Average Performance on  Neural 3D Video Dataset without Image Undistortion.} We use the best initial frame reconstruction from 10 runs for sequential reconstruction, same as the main experiments.}
    \begin{tabular}{l|ccc}
    \toprule
    \multirow{2}{*}{Model}  &   \multicolumn{3}{c}{Neural 3D Video}  \\
    & PSNR $\uparrow$ & SSIM $\uparrow$ & mTV$_{\times100}$ $\downarrow$ \\
    \midrule
   Dynamic3DG$\dagger$     & 32.12 & 0.952 & 0.169         \\ 
   Dynamic3DG$\dagger$ + ours  & \textbf{32.84} & \textbf{0.956} & \textbf{0.077}     \\ \cmidrule{1-4}
   3DGStream               & 32.11 & 0.952 & 0.276         \\
   3DGStream + ours        & \textbf{32.90} & \textbf{0.956} & \textbf{0.146}         \\ \cmidrule{1-4}
   HiCoM                   & 32.03 & 0.950 & 0.198         \\
   HiCoM + ours            & \textbf{32.13} & 0.950 & \textbf{0.129}         \\ 
    \bottomrule
    \end{tabular}
    \label{tab:n3v_dist}
\end{table}
\section{Additional Comparison}
\label{sup:offline}
\subsection{Experiments Details}
The experimental settings of HiCoM~\cite{gao2024hicom} are different from previous works~\cite{luiten2023dynamic,sun20243dgstream}. We conducted additional experiments using original images and MVS initialization, even though image undistortion improves the PSNR. We measured the mean and standard deviation across all 10 runs without selection, labeling as ``$\text{Online}^*$''. In this setting, we additionally compare the performance of online reconstruction methods and offline reconstruction methods.

\paragraph{Offline Reconstruction}
STG~\cite{li2023spacetimegaussians} originally reconstructs each video segment independently due to memory constraints, causing visual inconsistencies between segments. For a fair comparison, we modified the implementation of STG to train on the complete video, following 4D-Scaffold~\cite{cho20244dscaffold}.

\subsection{Comparison}
Applying our method improves both temporal consistency (reduced mTV) and visual quality (increased PSNR) as shown in \tref{tab:n3v_w_offline} and \tref{tab:n3v_dist}.
Moreover, our method demonstrates greater stability with lower standard deviation than baselines, as shown by smaller ellipse radii in the scatter plot (\fref{fig:scatter}).

\subsection{Comparison to Offline Reconstruction}
3DGStream with our methods achieves the best visual quality among online and offline methods. 
However, our method enhances the temporal consistency of online methods, it does not reach the consistency level of offline methods.
Our best results show 1.4$\times$ higher total variation compared to the worst offline reconstruction, thus it does not yet match offline methods. We attribute the gap to the nature of online configuration, which uses only a snapshot at a moment without the previous and next frames, leading to inevitable occlusions.

\begin{figure}[tb!]
    \centering
    \includegraphics[width=\linewidth]{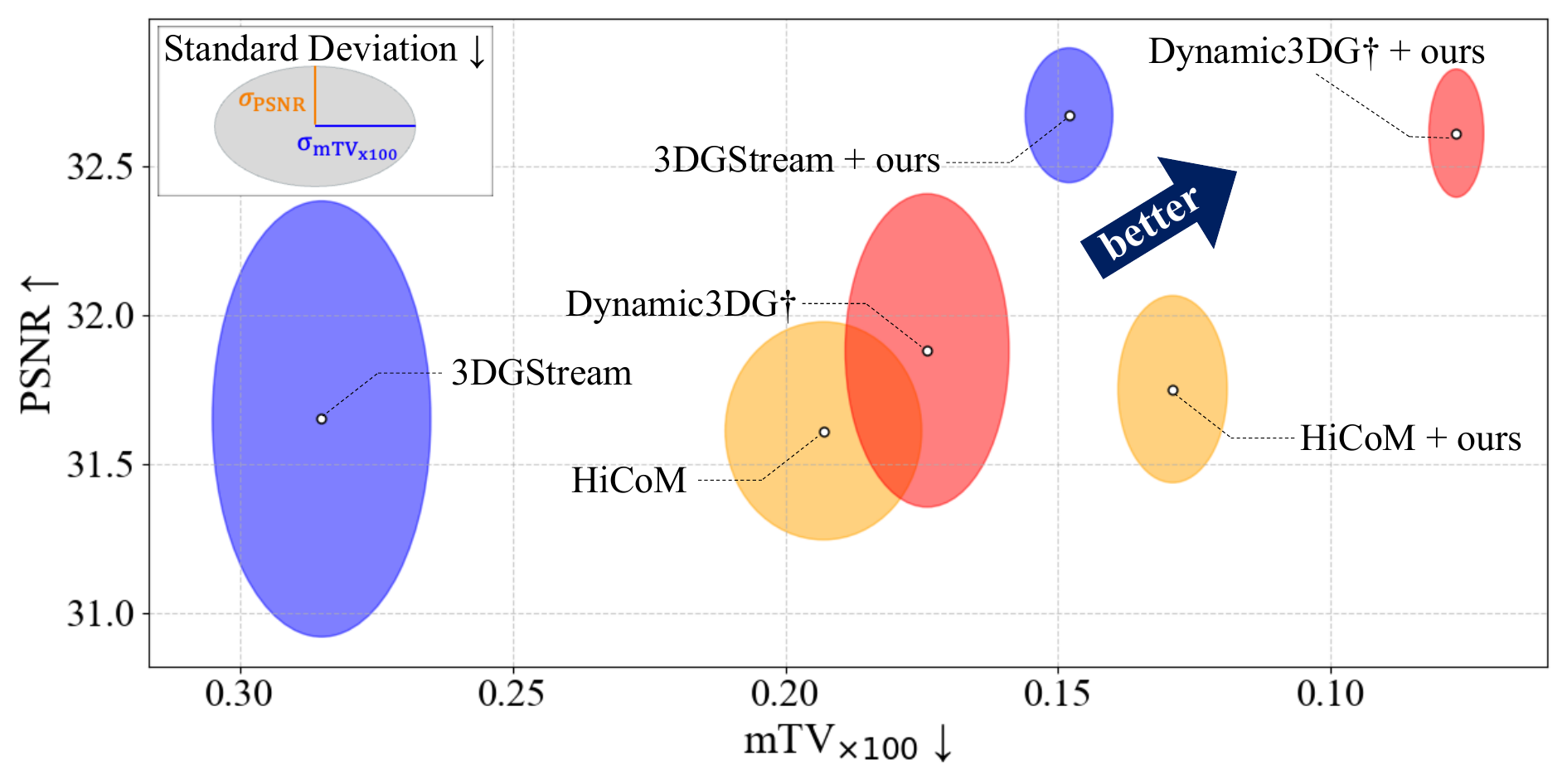}
    \caption{
    \textbf{Scatter Plots Across 10 Runs for each Scene in the Neural 3D Video Dataset.} The radii of the ellipse represent the standard deviations along each axis. 
    }
    \label{fig:scatter}
\Description[Figure]{scatter}
\end{figure}

\begin{figure}[tb!]
    \centering
    \includegraphics[width=\linewidth]{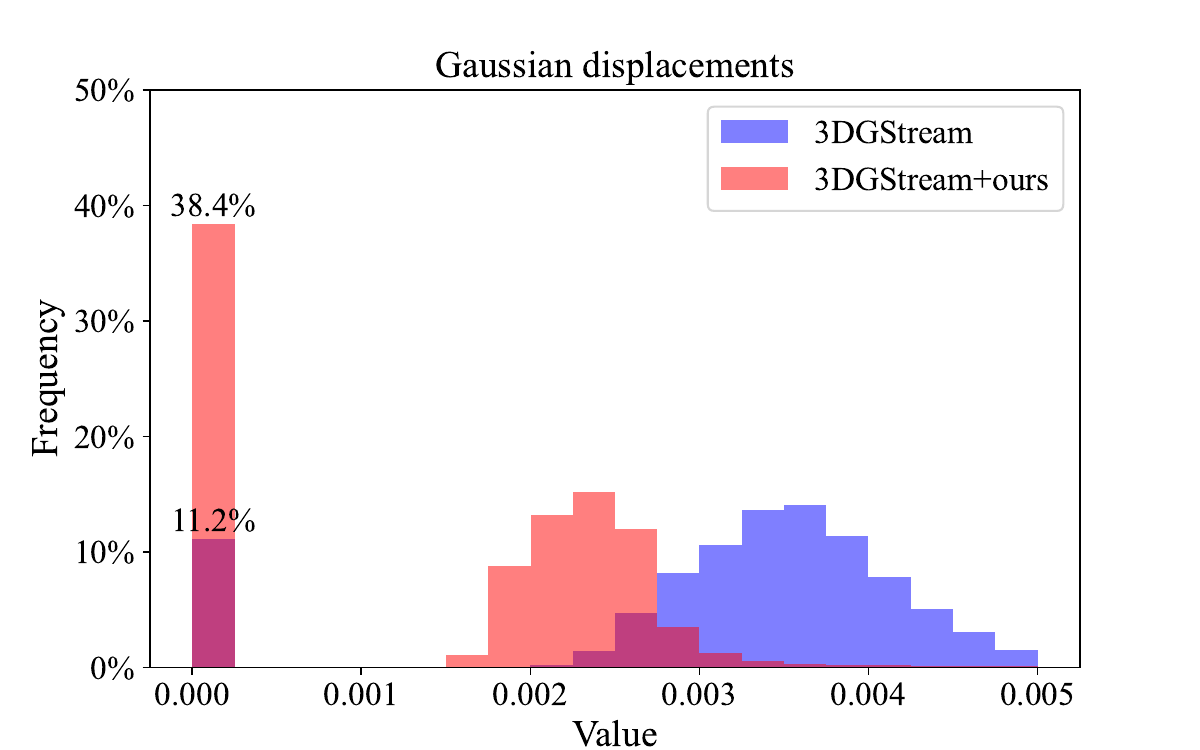}
    \caption{
    \textbf{Histogram of Gaussian Displacements in the \texttt{coffee\_martini} Scene.}
    }
    \label{fig:histogram}
\Description[Figure]{Histogram}
\end{figure}

\subsection{Comparison to V³}
V$^3$~\cite{wang2024v} proposes temporal entropy loss and temporal loss to enhance temporal continuity, leading to efficient compression. These losses are L2 regularization and entropy regularization on the scale, rotation, and opacity displacement of Gaussians\footnote{Unlike the original paper, their implementation does not apply loss to the color (https://github.com/AuthorityWang/VideoGS).}.

However, V$^3$ targets only the reconstruction of dynamic foregrounds and does not handle static backgrounds. Therefore, our method is orthogonal to V$^3$, which makes it unsuitable for demonstrating our effectiveness in enhancing temporal consistency in static regions. Moreover, V$^3$ applies the loss only within groups, without considering temporal continuity across groups, which makes it limited.

Therefore, instead of directly comparing with V$^3$, we compare against its proposed loss. In \tref{tab:v3}, our method outperforms loss of V$^3$, demonstrating superior effectiveness.

\begin{table}[tb!]
    \centering
    \caption{\textbf{Comparison to Temporal Loss on Neural 3D Video Dataset}.}
    \begin{tabular}{l|ccc}
        \toprule
        Model & PSNR $\uparrow$ & SSIM $\uparrow$ & mTV$_{\times100}$ $\downarrow$  \\
        \midrule
         3DGStream & 32.58 & 0.960 & 0.178 \\
         3DGStream + V$^3$ loss & 32.57 & 0.960 & 0.180 \\
         3DGStream + ours & \textbf{33.13} & \textbf{0.961} & \textbf{0.103} \\
        \bottomrule
    \end{tabular}
    \label{tab:v3}
\end{table}

\section{More Analysis}
\label{sup:more}
\subsection{Displacement Distribution}
\label{sup:more:histogram}
To demonstrate the movement of Gaussian over time, we compared 3DGStream with and without our method. \fref{fig:histogram} shows the histogram distribution of Gaussian position displacements. Our method increases the number of Gaussians with near-zero displacement by 3.4$\times$ compared to the baseline, indicating a substantial reduction in unnecessary motion of static Gaussians. This leads to more accurate scene representation and better temporal consistency in static regions.

\begin{figure}[tb!]
    \centering
    \includegraphics[width=0.875\linewidth]{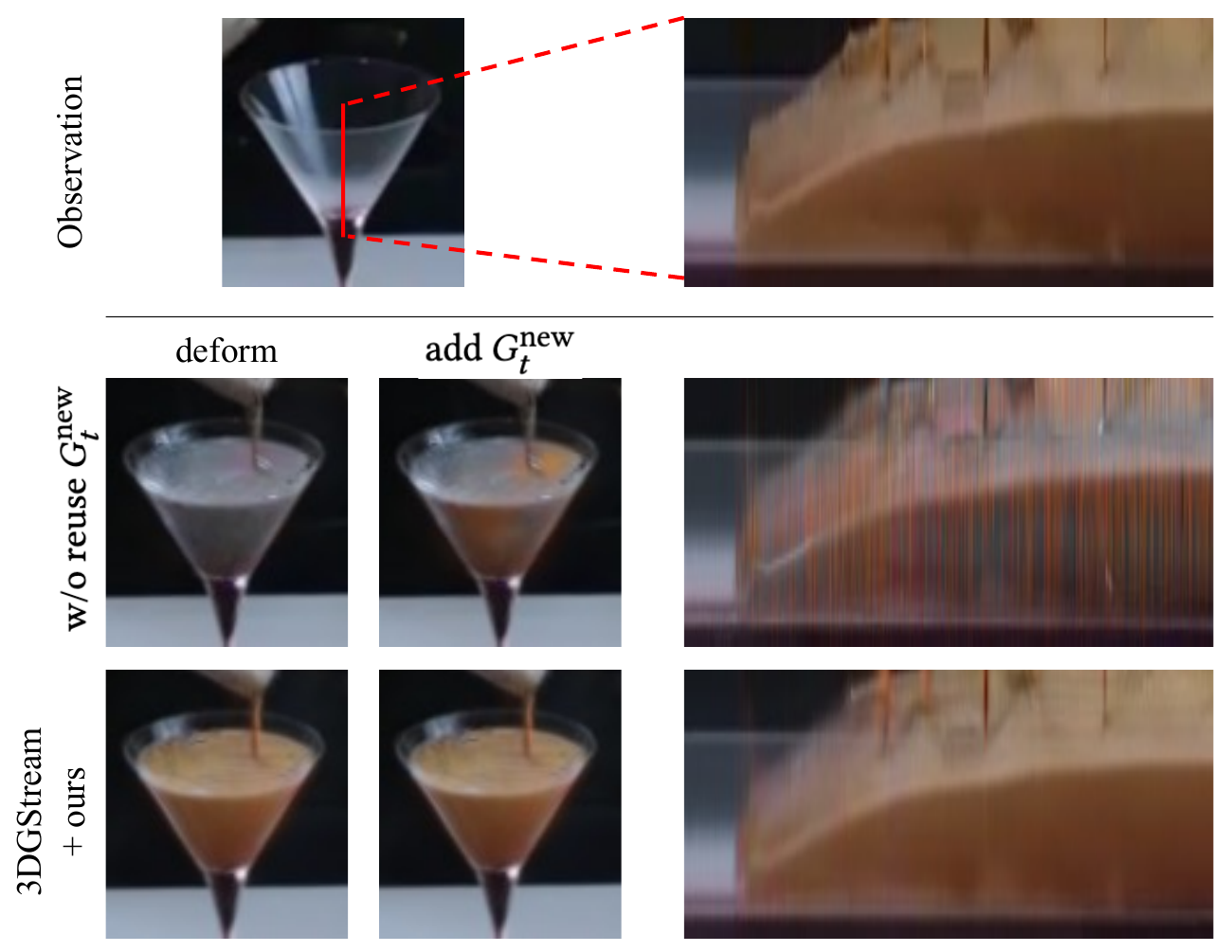}
    \caption{
        \textbf{Difference in Temporal Consistency with and without Reusing New Gaussians $G_t^\text{new}$ in Subsequent Frames.}
    }
    \label{fig:new_gaussian_consistency}
\Description[Figure]{Temporal consistency results}
\end{figure}

\begin{figure}[tb!]
    \centering
    \includegraphics[width=0.875\linewidth]{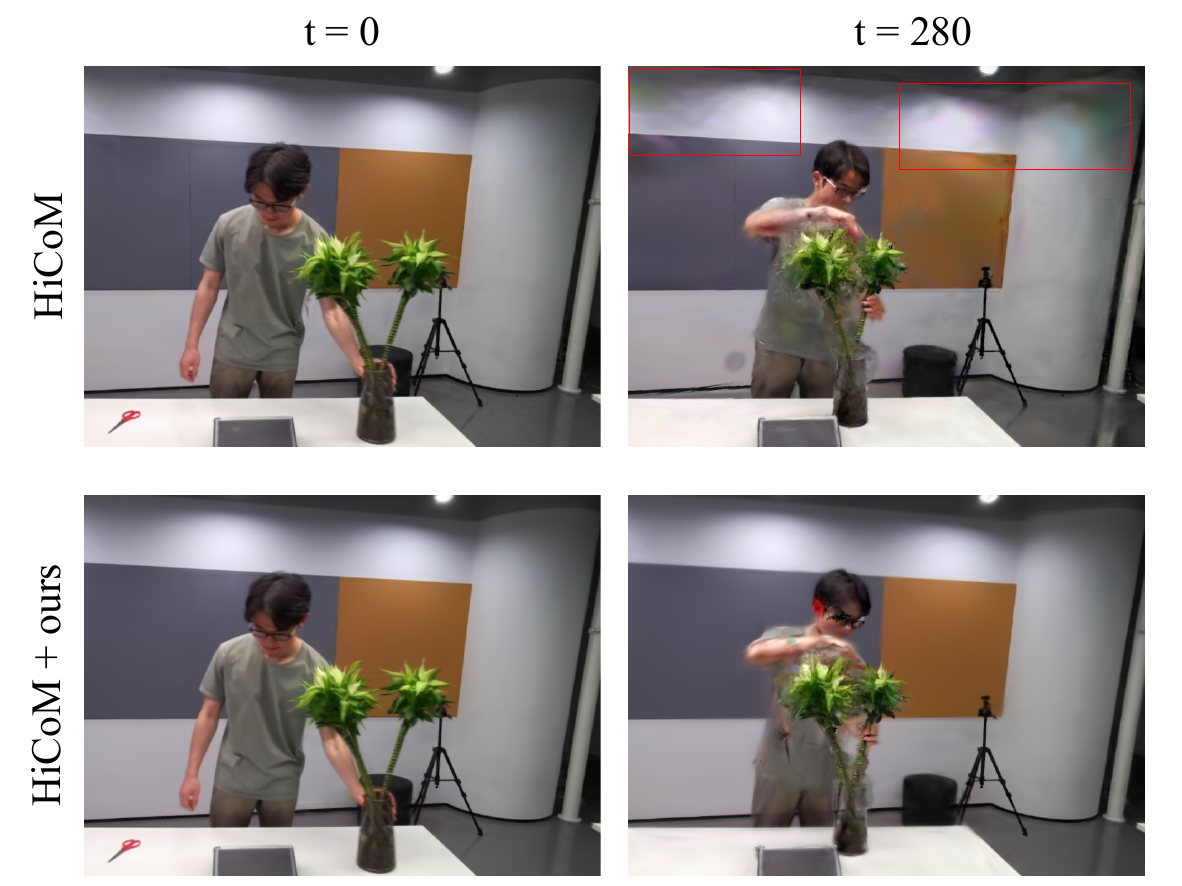}
    \caption{
    \textbf{HiCoM Exhibits Limited Static Region Preservation.}
    }
    \label{fig:hicom}
\Description[Figure]{Histogram}
\end{figure}

\subsection{New Gaussians and Temporal Consistency}
\label{sup:more:new}
We demonstrate that using new Gaussians only in a specific frame harms temporal consistency. As shown in \fref{fig:new_gaussian_consistency}, the coffee exhibits temporal inconsistency in ``w/o reuse $G_t^{\text{new}}$''. This occurs because Gaussians representing glass are deformed to represent coffee in subsequent frames, disrupting the optimization of new Gaussians. We show that this issue is resolved by propagating new Gaussians to subsequent frames in ``3DGStream+ours''.

\subsection{Additional Comparison to HiCoM}
Our experiments show that both the original HiCoM and HiCoM with our method fail to reconstruct fast motions as discussed in the limitation. However, \fref{fig:hicom} shows that our method performs notably better quality in static regions while the original HiCoM fails to preserve the quality of static regions.

\subsection{Various Noise Levels}
\label{sup:more:noise}
Noises in adjacent frames differ according to devices and environments. In our experiments, the observations are captured by various devices (GoPro and Kinect) and various environments. Their temporal inconsistency varies in [0.090,0.834].

As shown in \tref{tab:noise_level}, our method improves temporal consistency under various noise levels, though the improvement decreases with severe noise.

\begin{table}[tb!]
    \centering
    \caption{\textbf{Effect of Noise Levels on Temporal Inconsistency}. We report mTV$_{\times100}$ on Sync-NeRF dataset.}
    \resizebox{\linewidth}{!}{
    \begin{tabular}{l|ccc}
        \toprule
        Noise level (mTV$_{\times100}$) & 0.452 & 0.908 & 1.840  \\
        \midrule
         3DGStream & 0.483 & 0.559 & 0.887 \\
         3DGStream + ours & 0.318 (0.66$\times$) & 0.452 (0.81$\times$) & 0.815 (0.92$\times$) \\
        \bottomrule
    \end{tabular}
    }
    \label{tab:noise_level}
\end{table}

\subsection{Various Number of Train Views}
\label{sup:more:view}

\begin{table}[tb!]
    \centering
    \caption{\textbf{Effect of Number of Train Views on Temporal Inconsistency}. We report mTV$_{\times100}$ on Sync-NeRF dataset.}
    \resizebox{\linewidth}{!}{
    \begin{tabular}{l|ccc}
        \toprule
        Number of views & 13 & 9 & 5  \\
        \midrule
         3DGStream & 0.483 & 0.764 & 0.924 \\
         3DGStream + ours & 0.318 (0.66$\times$) & 0.645 (0.84$\times$) & 0.829 (0.90$\times$) \\
        \bottomrule
    \end{tabular}
    }
    \label{tab:number_of_views}
\end{table}

In \tref{tab:number_of_views}, we validate our method under the number of training views. While our method improves temporal consistency across various number of training views, its effectiveness decreases as the number of views is reduced, since it relies on well-reconstructed scenes and multi-view consistency of Gaussians to fit multi-view inconsistent noise.

\section{More Results}
We report the quantitative results of each scene in \tref{tab:per_scene}. Our method enhances temporal consistency across all scenes (reduced mTV). We report per-frame metrics in \fref{fig:per-frame}, where our method outperforms the baseline in most cases.

\begin{table*}[bht]
    \centering
    \caption{\textbf{Per-scene Quantitative Results on Neural 3D Video and Meetroom Dataset.} {We report mean {\scriptsize $\pm$ standard deviation} of video frames.}}
    \resizebox{\linewidth}{!}{
    \begin{tabular}{l|ccc|ccc|ccc}
    \toprule
    \multirow{2}{*}{Model} & \multicolumn{3}{c|}{\texttt{coffee\_martini}} & \multicolumn{3}{c|}{\texttt{cook\_spinach}} & \multicolumn{3}{c}{\texttt{cut\_roasted\_beef}} \\
         & PSNR $\uparrow$ & SSIM $\uparrow$ & mTV$_{\times100}$ $\downarrow$ & PSNR $\uparrow$ & SSIM $\uparrow$ & mTV$_{\times100}$ $\downarrow$ & PSNR $\uparrow$ & SSIM $\uparrow$ & mTV$_{\times100}$ $\downarrow$ \\
         \midrule
         Dynamic3DG$\dagger$ & 29.36 {\scriptsize $\pm$ 0.06} & 0.938 {\scriptsize $\pm$ 0.001} & 0.169 {\scriptsize $\pm$ 0.027} & 33.86 {\scriptsize $\pm$ 0.33} & \textbf{0.968} {\scriptsize $\pm$ 0.001} & 0.127 {\scriptsize $\pm$ 0.018} & 33.67 {\scriptsize $\pm$ 0.67} & 0.969 {\scriptsize $\pm$ 0.003} & 0.226 {\scriptsize $\pm$ 0.050}     \\
         Dynamic3DG$\dagger$ + ours & \textbf{29.38} {\scriptsize $\pm$ 0.06} & \textbf{0.939} {\scriptsize $\pm$ 0.001} & \textbf{0.057} {\scriptsize $\pm$ 0.071} & \textbf{34.34} {\scriptsize $\pm$ 0.41} & 0.966 {\scriptsize $\pm$ 0.002} & \textbf{0.048} {\scriptsize $\pm$ 0.034} & \textbf{34.66} {\scriptsize $\pm$ 0.36} & 0.969 {\scriptsize $\pm$ 0.002} & \textbf{0.059} {\scriptsize $\pm$ 0.035}     \\ \midrule
         3DGStream           & 29.24 {\scriptsize $\pm$ 0.09} & 0.938 {\scriptsize $\pm$ 0.001} & 0.163 {\scriptsize $\pm$ 0.034} & 33.96 {\scriptsize $\pm$ 0.36} & 0.966 {\scriptsize $\pm$ 0.002} & 0.179 {\scriptsize $\pm$ 0.032} & 33.99 {\scriptsize $\pm$ 0.19} & 0.968 {\scriptsize $\pm$ 0.002} & 0.165 {\scriptsize $\pm$ 0.039}     \\
         3DGStream + ours    & \textbf{29.54} {\scriptsize $\pm$ 0.06} & \textbf{0.941} {\scriptsize $\pm$ 0.001} & \textbf{0.082} {\scriptsize $\pm$ 0.070} & \textbf{34.52} {\scriptsize $\pm$ 0.38} & \textbf{0.967} {\scriptsize $\pm$ 0.002} & \textbf{0.105} {\scriptsize $\pm$ 0.035} & \textbf{34.78} {\scriptsize $\pm$ 0.19} & 0.968 {\scriptsize $\pm$ 0.002} & \textbf{0.097} {\scriptsize $\pm$ 0.039}     \\ \midrule
         HiCoM               & 29.14 {\scriptsize $\pm$ 0.09} & 0.936 {\scriptsize $\pm$ 0.001} & 0.256 {\scriptsize $\pm$ 0.026} & 33.39 {\scriptsize $\pm$ 0.50} & 0.961 {\scriptsize $\pm$ 0.003} & 0.140 {\scriptsize $\pm$ 0.023} & 34.00 {\scriptsize $\pm$ 0.27} & \textbf{0.965} {\scriptsize $\pm$ 0.003} & 0.119 {\scriptsize $\pm$ 0.027}     \\
         HiCoM + ours        & \textbf{29.26} {\scriptsize $\pm$ 0.12} & \textbf{0.937} {\scriptsize $\pm$ 0.002} & \textbf{0.159} {\scriptsize $\pm$ 0.018} & \textbf{33.80} {\scriptsize $\pm$ 0.49} & \textbf{0.962} {\scriptsize $\pm$ 0.003} & \textbf{0.083} {\scriptsize $\pm$ 0.012} & \textbf{34.02} {\scriptsize $\pm$ 0.39} & 0.964 {\scriptsize $\pm$ 0.003} & \textbf{0.089} {\scriptsize $\pm$ 0.014}     \\ \midrule\midrule
          \multirow{2}{*}{Model} & \multicolumn{3}{c|}{\texttt{flame\_salmon}} & \multicolumn{3}{c|}{\texttt{flame\_steak}} & \multicolumn{3}{c}{\texttt{sear\_steak}} \\
         & PSNR $\uparrow$ & SSIM $\uparrow$ & mTV$_{\times100}$ $\downarrow$ & PSNR $\uparrow$ & SSIM $\uparrow$ & mTV$_{\times100}$ $\downarrow$ & PSNR $\uparrow$ & SSIM $\uparrow$ & mTV$_{\times100}$ $\downarrow$ \\
         \midrule
         Dynamic3DG$\dagger$ & 29.70 {\scriptsize $\pm$ 0.13} & 0.940 {\scriptsize $\pm$ 0.001} & 0.145 {\scriptsize $\pm$ 0.028} & 33.88 {\scriptsize $\pm$ 0.54} & 0.973 {\scriptsize $\pm$ 0.001} & 0.123 {\scriptsize $\pm$ 0.027} & 34.42 {\scriptsize $\pm$ 0.15} & 0.974 {\scriptsize $\pm$ 0.001} & 0.126 {\scriptsize $\pm$ 0.018}     \\
         Dynamic3DG$\dagger$ + ours & \textbf{29.75} {\scriptsize $\pm$ 0.16} & \textbf{0.942} {\scriptsize $\pm$ 0.001} & \textbf{0.053} {\scriptsize $\pm$ 0.062} & \textbf{35.03} {\scriptsize $\pm$ 0.45} & 0.973 {\scriptsize $\pm$ 0.001} & \textbf{0.051} {\scriptsize $\pm$ 0.046} & \textbf{35.01} {\scriptsize $\pm$ 0.24} & 0.974 {\scriptsize $\pm$ 0.001} & \textbf{0.051} {\scriptsize $\pm$ 0.033}     \\ \midrule
         3DGStream           & 29.67 {\scriptsize $\pm$ 0.13} & 0.941 {\scriptsize $\pm$ 0.001} & 0.205 {\scriptsize $\pm$ 0.046} & 34.29 {\scriptsize $\pm$ 0.43} & 0.973 {\scriptsize $\pm$ 0.001} & 0.177 {\scriptsize $\pm$ 0.045} & 34.34 {\scriptsize $\pm$ 0.26} & 0.973 {\scriptsize $\pm$ 0.001} & 0.180 {\scriptsize $\pm$ 0.040}     \\
         3DGStream + ours    & \textbf{29.99} {\scriptsize $\pm$ 0.11} & \textbf{0.944} {\scriptsize $\pm$ 0.001} & \textbf{0.116} {\scriptsize $\pm$ 0.061} & \textbf{34.86} {\scriptsize $\pm$ 0.37} & 0.973 {\scriptsize $\pm$ 0.001} & \textbf{0.103} {\scriptsize $\pm$ 0.046} & \textbf{35.11} {\scriptsize $\pm$ 0.20} & \textbf{0.974} {\scriptsize $\pm$ 0.001} & \textbf{0.113} {\scriptsize $\pm$ 0.036}     \\ \midrule
         HiCoM               & 29.46 {\scriptsize $\pm$ 0.26} & 0.939 {\scriptsize $\pm$ 0.003} & 0.218 {\scriptsize $\pm$ 0.056} & 32.48 {\scriptsize $\pm$ 0.37} & 0.968 {\scriptsize $\pm$ 0.001} & 0.111 {\scriptsize $\pm$ 0.014} & 34.08 {\scriptsize $\pm$ 0.22} & \textbf{0.971} {\scriptsize $\pm$ 0.001} & 0.137 {\scriptsize $\pm$ 0.019}     \\
         HiCoM + ours        & \textbf{29.59} {\scriptsize $\pm$ 0.65} & 0.939 {\scriptsize $\pm$ 0.007} & \textbf{0.153} {\scriptsize $\pm$ 0.065} & \textbf{34.07} {\scriptsize $\pm$ 0.61} & \textbf{0.969} {\scriptsize $\pm$ 0.002} & \textbf{0.091} {\scriptsize $\pm$ 0.014} & \textbf{34.44} {\scriptsize $\pm$ 0.37} & 0.970 {\scriptsize $\pm$ 0.002} & \textbf{0.100} {\scriptsize $\pm$ 0.016}     \\ \midrule\midrule
         \multirow{2}{*}{Model} & \multicolumn{3}{c|}{\texttt{discussion}} & \multicolumn{3}{c|}{\texttt{trimming}} & \multicolumn{3}{c}{\texttt{vrheadset}} \\
         & PSNR $\uparrow$ & SSIM $\uparrow$ & mTV$_{\times100}$ $\downarrow$ & PSNR $\uparrow$ & SSIM $\uparrow$ & mTV$_{\times100}$ $\downarrow$ & PSNR $\uparrow$ & SSIM $\uparrow$ & mTV$_{\times100}$ $\downarrow$ \\
         \midrule
         Dynamic3DG$\dagger$ & 31.32 {\scriptsize $\pm$ 0.36} & 0.956 {\scriptsize $\pm$ 0.002} & 0.192 {\scriptsize $\pm$ 0.019} & \textbf{31.73} {\scriptsize $\pm$ 0.19} & \textbf{0.953} {\scriptsize $\pm$ 0.001} & 0.173 {\scriptsize $\pm$ 0.011} & 29.32 {\scriptsize $\pm$ 0.95} & 0.948 {\scriptsize $\pm$ 0.003} & 0.158 {\scriptsize $\pm$ 0.011}     \\
         Dynamic3DG$\dagger$ + ours & \textbf{31.95} {\scriptsize $\pm$ 0.96} & \textbf{0.959} {\scriptsize $\pm$ 0.003} & \textbf{0.145} {\scriptsize $\pm$ 0.137} & 31.66 {\scriptsize $\pm$ 0.25} & 0.952 {\scriptsize $\pm$ 0.002} & \textbf{0.127} {\scriptsize $\pm$ 0.037} & \textbf{30.27} {\scriptsize $\pm$ 1.13} & \textbf{0.949} {\scriptsize $\pm$ 0.004} & \textbf{0.107} {\scriptsize $\pm$ 0.031}     \\ \midrule
         3DGStream           & 31.78 {\scriptsize $\pm$ 0.30} & 0.956 {\scriptsize $\pm$ 0.003} & 0.055 {\scriptsize $\pm$0.009 } & 32.29 {\scriptsize $\pm$ 0.21} & 0.957 {\scriptsize $\pm$ 0.001} & 0.039 {\scriptsize $\pm$ 0.008} & 31.14 {\scriptsize $\pm$ 0.19} & 0.953 {\scriptsize $\pm$ 0.001} & 0.052 {\scriptsize $\pm$ 0.001}     \\ 
         3DGStream + ours    & \textbf{32.51} {\scriptsize $\pm$ 0.36} & \textbf{0.958} {\scriptsize $\pm$ 0.002} & \textbf{0.027} {\scriptsize $\pm$ 0.004} & \textbf{32.78} {\scriptsize $\pm$ 0.20} & \textbf{0.958} {\scriptsize $\pm$ 0.001} & \textbf{0.018} {\scriptsize $\pm$ 0.004} & \textbf{31.58} {\scriptsize $\pm$ 0.35} & 0.953 {\scriptsize $\pm$ 0.002} & \textbf{0.027} {\scriptsize $\pm$ 0.000}     \\ \midrule
         HiCoM               & \textbf{29.02} {\scriptsize $\pm$ 1.56} & 0.929 {\scriptsize $\pm$ 0.017} & 0.166 {\scriptsize $\pm$ 0.075} & 29.48 {\scriptsize $\pm$ 1.14} & 0.936 {\scriptsize $\pm$ 0.011} & 0.131 {\scriptsize $\pm$ 0.070} & \textbf{29.75} {\scriptsize $\pm$ 0.67} & \textbf{0.943} {\scriptsize $\pm$ 0.004} & 0.083 {\scriptsize $\pm$ 0.014}     \\
         HiCoM + ours        & 28.87 {\scriptsize $\pm$ 0.73} & \textbf{0.931} {\scriptsize $\pm$ 0.008} & \textbf{0.056} {\scriptsize $\pm$ 0.011} & \textbf{29.69} {\scriptsize $\pm$ 0.82} & \textbf{0.940} {\scriptsize $\pm$ 0.006} & \textbf{0.044} {\scriptsize $\pm$ 0.015} & 29.71 {\scriptsize $\pm$ 0.50} & 0.941 {\scriptsize $\pm$ 0.003} & \textbf{0.047} {\scriptsize $\pm$ 0.005}     \\ 
    \bottomrule
    \end{tabular}
    }
    \label{tab:per_scene}
\end{table*}

\begin{figure*}[tb!]
    \centering
    \includegraphics[width=\linewidth]{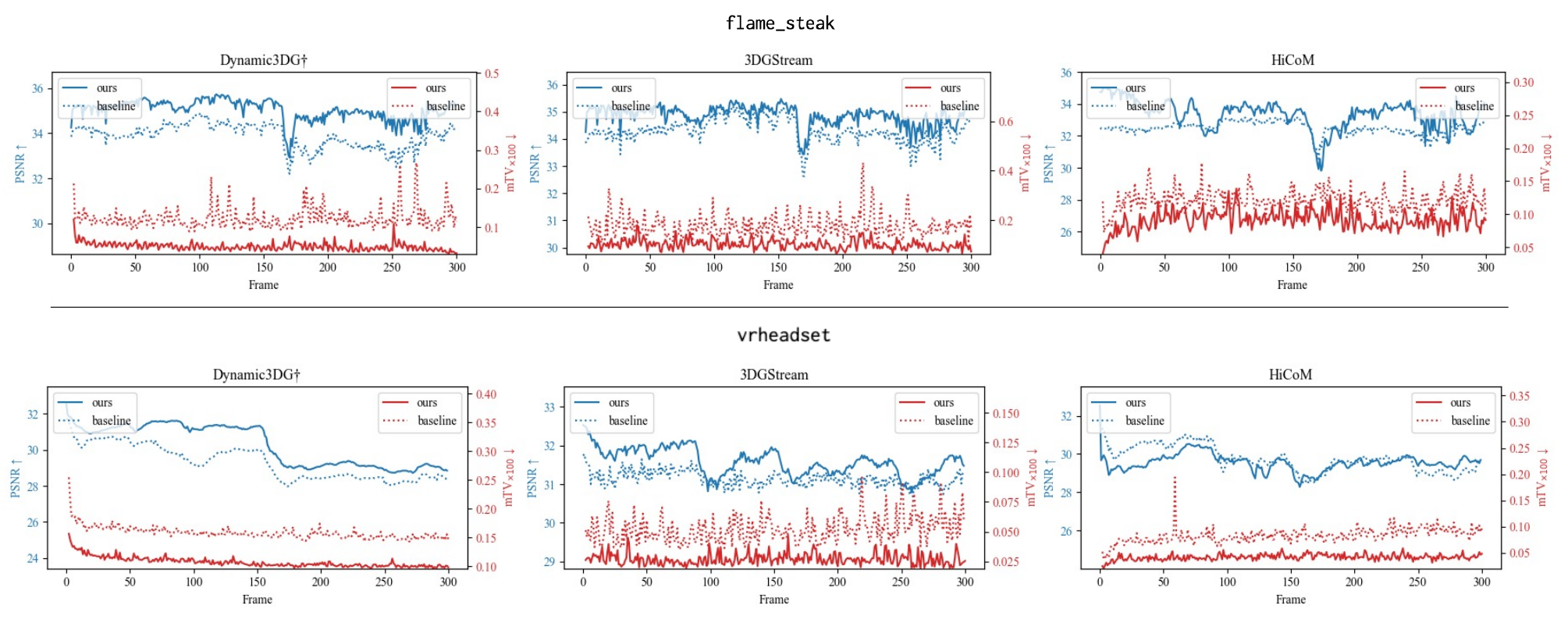}
    \caption{
    \textbf{Per-frame Metrics of \texttt{flame\_steak} and \texttt{vrheadset} Scene.}
    }
    \label{fig:per-frame}
\Description[Figure]{Per-frame}
\end{figure*}
\fi

\end{document}